\documentclass[final,5p,times,twocolumn]{elsarticle}

\usepackage{amssymb}
\usepackage{amsmath}
\usepackage{url}
\usepackage{verbatim}
\usepackage{hyperref}
\usepackage{float}

\hypersetup{
    colorlinks=true,       
    linkcolor=blue,        
    citecolor=blue,         
    filecolor=magenta,     
    urlcolor=teal          
}

\journal{}

\begin{document}



\title{Trajectory-based road autolabeling with lidar-camera fusion in winter conditions}


\author[a]{Eerik Alamikkotervo}
\ead{eerik.alamikkotervo@aalto.fi}
\author[a]{Henrik Toikka}
\ead{henrik.toikka@aalto.fi}
\author[a]{Kari Tammi}
\ead{kari.tammi@aalto.fi}
\author[a]{Risto Ojala}
\ead{risto.j.ojala@aalto.fi}

\affiliation[a]{Aalto University, Otakaari 4, Espoo, 02150, Uusimaa, Finland}

\begin{abstract}
Robust road segmentation in all road conditions is required for safe autonomous driving and advanced driver assistance systems.
Supervised deep learning methods provide accurate road segmentation in the domain of their training data but cannot be trusted in out-of-distribution scenarios. 
Including the whole distribution in the trainset is challenging as each sample must be labeled by hand. 
Trajectory-based self-supervised methods offer a potential solution as they can learn from the traversed route without manual labels. 
However, existing trajectory-based methods use learning schemes that rely only on the camera or only on the lidar.  
In this paper, trajectory-based learning is implemented jointly with lidar and camera for increased performance. 
Our method outperforms recent standalone camera- and lidar-based methods when evaluated with a challenging winter driving dataset including countryside and suburb driving scenes. The source code is available at \url{https://github.com/eerik98/lidar-camera-road-autolabeling.git}. The winter driving dataset is made publically available at  \url{https://zenodo.org/records/14856338}. 
\end{abstract}



\maketitle

\section{Introduction}

Robust road segmentation is essential for autonomous driving and Advanced Driver Assistance Systems (ADAS). 
Supervised deep learning approaches can achieve high road segmentation accuracy by learning complex detection patterns from large manually labeled datasets \cite{geiger2013vision,caesar2020nuscenes,yu2020bdd100k,cordts2016cityscapes}. 
However, these datasets require significant time and resources to create and the models are only reliable in the domain of their training data \cite{shaik2024idd,sakaridis2021acdc}, making them unpractical to use in diverse driving scenarios. 

Trajectory-based learning has been proposed as an alternative to avoid the need for manual labeling \cite{schmid2022self, seo2023learning, jung2024v}. 
These methods aim to learn the appearance of the road based on the traversed route. 
However, accurately estimating traversability solely from visual appearance remains challenging, especially in complex environments.  
Incorporating depth information from sensors like stereo cameras or lidar could increase the performance significantly.  

In this paper, we propose a novel trajectory-based autolabeling method leveraging both camera and lidar. We are the first to propose lidar-camera fusion for trajectory-based learning. Our method is validated with a diverse winter driving dataset,  demonstrating significant improvements over existing self-supervised methods and even surpassing a supervised baseline trained with manually labeled data. The main contributions are:

\begin{itemize}
    \item Utilizing lidar and camera jointly for trajectory-based learning. Existing methods rely only on lidar \cite{seo2023scate} or camera \cite{seo2023learning, schmid2022self, jung2024v, alamikkotervo2024tadap}.

    \item Validation in challenging winter driving conditions. Existing works validated in mostly non-snowy roads. The winter driving dataset is made publically available for future development.  

    \item We outperform both lidar and camera-based baselines with a clear margin. Source code is made publically available for future development. To the best of our knowledge, we are the only trajectory-based method with public source code. 

\end{itemize}

\section{Related work}

\subsection{Trajectory-based learning}
Trajectory-based learning utilizes the traversed route as the only supervision. 
Vehicle wheel trajectories are estimated based on the recorded poses and the area between the wheels is defined as a partial positive label. 
Trajectory-based learning methods are trained to predict the whole traversable area using the partial label as the only supervision. During prediction, the partial label is not available. 
Most recent trajectory-based methods use lidar-based localization for estimating the wheel trajectories, but the rest of the method is fully camera-based \cite{seo2023learning,schmid2022self,jung2024v}. 
Others rely fully on lidar and do not use camera at all \cite{seo2023scate}. 
It is challenging to segment the whole traversable area based on a small trajectory sample while relying on a single sensor. 
Arguably, performance gains could be achieved by using lidar and camera jointly. 

\subsection{Lidar-camera fusion for traversability estimation}
Lidar-camera fusion is commonly used for traversability estimation, but none of the fusion methods are trajectory-based. 
Hand-crafted image and lidar features have been combined for traversability estimation in structured \cite{liu2018co} and off-road environments \cite{sock2016probabilistic}. 
However, hand-crafted image features are not robust in complex environments with varying road appearance and illumination.  
In \cite{frey2024roadrunner} a pre-trained deep-learning-based terrain segmenter is used instead, but the terrain segmenter can't adapt while driving as it is pre-trained. 
Lidar can also be used to automatically generate traversability labels for camera-based detection \cite{chen2023learning,wang2019self,mayr2018self,ma2023self}. Here lidar is simply used to generate ground truth and is not used for inference. 
Lidar-camera fusion is also used with supervised deep-learning models to increase the robustness of road detection in adverse conditions where lidar and camera suffer from weather effects \cite{rawashdeh2023camera}. 

\subsection{Vision foundation models}
Vision foundation models, pre-trained with large datasets, yield excellent zero-shot performance even in unseen environments. 
DINO \cite{caron2021emerging} and DINOv2 \cite{oquab2023dinov2}  utilize a self-supervised learning scheme to provide high-level feature representation for images. 
These features can be utilized for downstream tasks like segmentation, classification, and depth estimation by training a small model on top of the output features. 
On the other hand,  SAM \cite{kirillov2023segment} and SAM2 \cite{ravi2024sam} are trained with a large segmentation dataset for accurate zero-shot segmentation out of the box. 
Segmentation labels are created around query points provided by the user. 

\subsection{Lidar road boundary detection}
Lidars provide accurate 3D information of the environment which allows road boundary detection with simple handcrafted features. 
The scan is usually divided into scan rings and points are processed in order based on the polar angle. 
With this structure, effective features can be defined based on the height profile, height gradients, and scan ring deformations that occur on the road boundaries \cite{sun20193d,wang2020speed,zhang2018road}.
However, these methods tend to rely on manually tuned parameters and they are only reliable in scenarios where the road boundary is clearly defined.

\subsection{Research Gap}
In this work, we present the first trajectory-based approach utilizing both lidar and camera. 
Our camera-based detection leverages the vision foundation model DinoV2 to evaluate the visual similarity with the trajectory as demonstrated in our previous work \cite{alamikkotervo2024tadap}. 
Others \cite{jung2024v} have used SAM \cite{kirillov2023segment}, but it directly outputs segmentation masks that tend to be inaccurate in unstructured environments. 
Our lidar-based detection incorporates principles of lidar boundary detection with a trajectory-based learning approach. 
Traditional lidar boundary detection techniques rely on tunable parameters and struggle in unstructured environments \cite{sun20193d,zhang2018road,wang2020speed}. 
To address this issue, our method identifies trajectory points at each scan ring and uses them as reference points when evaluating the points at that ring. 

\section{Methods}
\begin{figure}
    \centering
    \includegraphics[width=\linewidth]{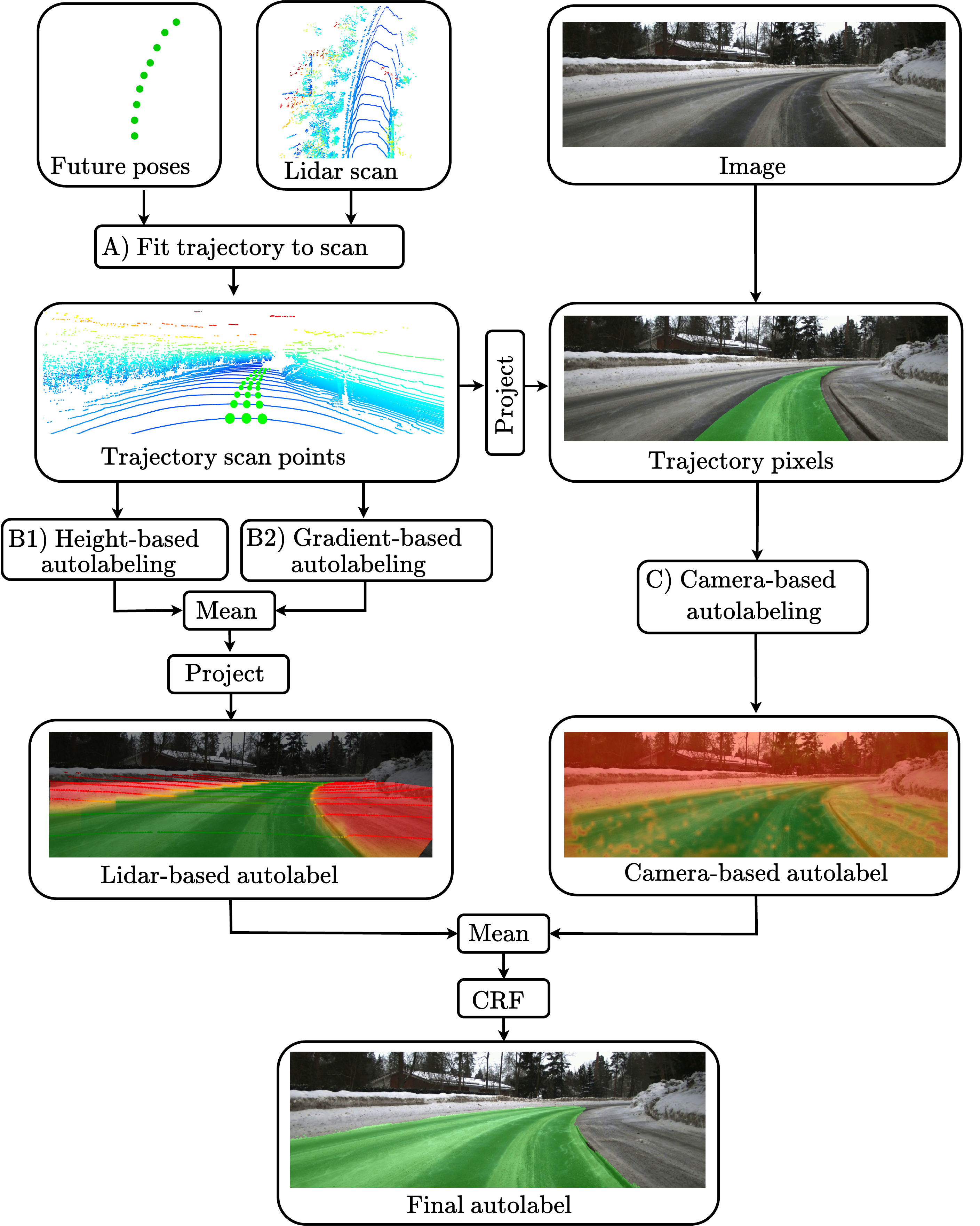}
    \caption{Proposed trajectory-based autolabeling with lidar-camera fusion. 
    Future poses are matched to the lidar scan to define trajectory scan points, which are then projected to the image frame to define trajectory pixels. 
    Lidar autolabeling uses height and gradient data, while camera autolabeling uses feature similarity. 
    The mean of the camera- and lidar-based labels is post-processed with CRF to yield the final label.}
    \label{fig:architecture}
\end{figure}

In this section, our novel trajectory-based autolabeling method utilizing both lidar and camera is presented (Fig. \ref{fig:architecture}). 
Our method first finds the trajectory points in the lidar scan (Section \ref{sec:match}).  
The lidar wheel points are projected to the image frame and all pixels inside the polygon defined by them are considered as trajectory pixels. 
Using the lidar trajectory points as a reference a lidar-based autolabel $l^{\text{lid}}$ is computed (Section \ref{sec:lidar}), and using the trajectory pixels as a reference a camera-based autolabel $l^{\text{cam}}$ is computed (Section \ref{sec:camera}). 
Both autolabels are continuous ranging from 0 to 1, with 1 indicating a definite road and 0 indicating a definite background.
The fused label is the mean of the camera-based and lidar-based autolabel.
Beyond lidar's sensing range we fully rely on the camera-based label.
Finally, the fused label is refined to discrete form with Conditional Random Field (CRF) post-processing \cite{krahenbuhl2011efficient}. 

\subsection{Fit trajectory to scan}
\label{sec:match}

\begin{figure}
    \centering
    \includegraphics[width=\linewidth]{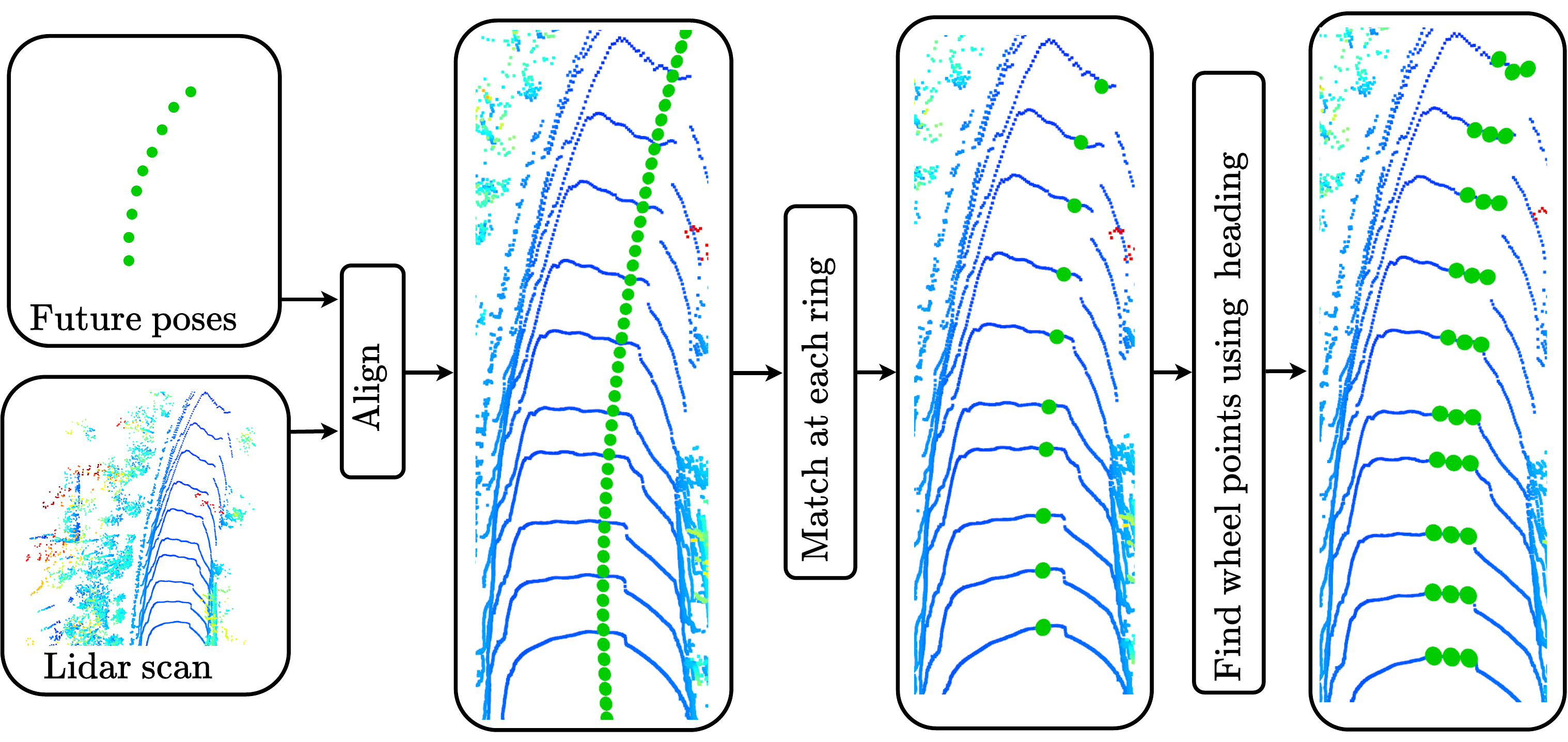}
    \caption{Finding the trajectory points from the scan. First future poses and the scan are aligned to the same frame. Then the closest match is found on each scan ring. Finally, wheel points are extended to each side based on the heading.}
    \label{fig:fit_trajectory}
\end{figure}

The objective is to identify lidar points corresponding to the center, the left wheel, and the right wheel of the vehicle at each scan ring, using recorded vehicle poses (Fig. \ref{fig:fit_trajectory}).
To locate the center point, we identify the lidar point nearest to the recorded poses within each scan ring. 
Based on the vehicle’s heading and known track width, the estimated positions of the left and right wheels are extended outward from the center point and the lidar points closest to these estimated wheel positions are selected. 

\begin{figure}
    \centering
    \setlength{\tabcolsep}{0pt}  
    \begin{tabular}{cc}
        \includegraphics[width=0.5\linewidth]{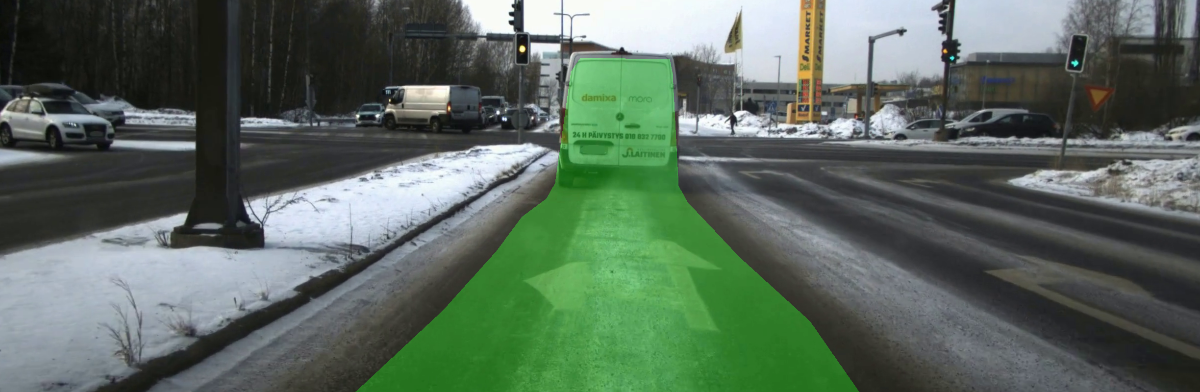} & 
        \includegraphics[width=0.5\linewidth]{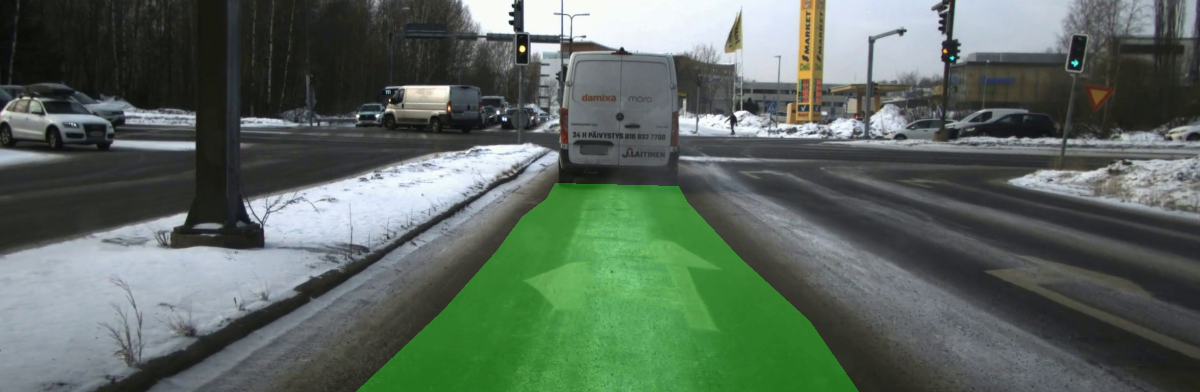} \\
        \includegraphics[width=0.5\linewidth]{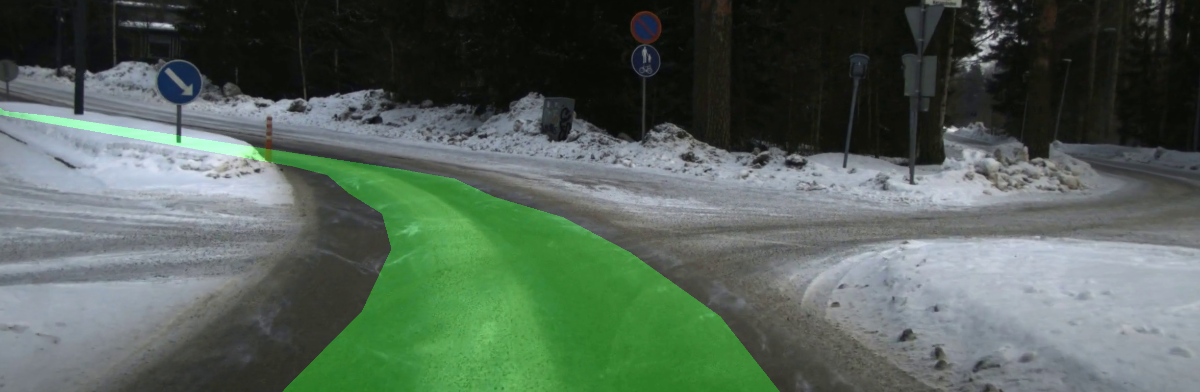} & 
        \includegraphics[width=0.5\linewidth]{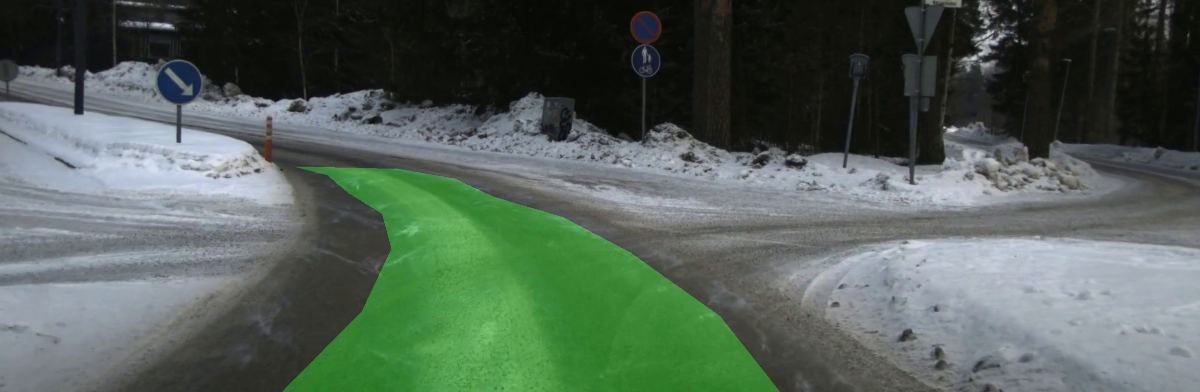} \\
    \end{tabular}
    \caption{Top row: False center points rejected based on rapid elevation increase caused by the van. Bottom row: occluded wheel points are rejected.}
    \label{fig:trajectory_filtering}
\end{figure}

The lidar scan is limited to a 90-degree field of view as we only need to consider points visible in the camera and the following filtering steps are applied to avoid false detections. 
The distance between the pose and the selected scan point must be less than 1 m, the distance between consecutive center points must be more than 1 m and the elevation difference between consecutive center points must be less than 1 m (Fig. \ref{fig:trajectory_filtering}). The wheel point distance from the center point must be less than 2 m and it must not be occluded (Fig. \ref{fig:trajectory_filtering}). Occlusion filtering is automatic. First lidar scan points (x,y,z) are converted to pixel coordinates (u,v).  We consider the line of sight to wheel point $p_{wheel}$ blocked by point $p$ if $|{u_{p}-u_{p_{wheel}}}|<10 \text{ pixels} $ and $v_{p}<v_{p_{wheel}}$. If either of the wheel points or the center points is filtered out that scan ring is discarded as we don't have all required reference points. The rest of the scan rings are still used. The parameter values proposed here were found to work well with our 32-channel lidar.

\subsection{Lidar-based autolabeling}
\label{sec:lidar}
The lidar autolabel $l^{\text{lid}}$ is the mean of the height-based label $l^{\text{h}}$ and gradient-based label $l^{\text{grad}}$ that are computed based on the known trajectory points at each scan ring including vehicle center, left wheel, and right wheel. 
The pointwise label is projected to the image frame and the gaps between the points are interpolated to produce a pixel-wise label. 

\subsubsection{Height-based autolabeling}
\begin{figure}
    \centering
    \includegraphics[width=\linewidth]{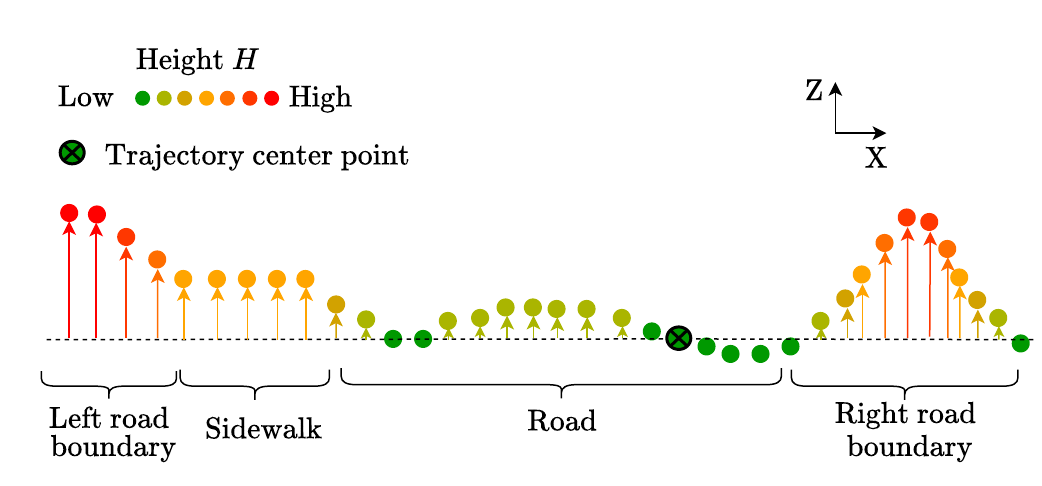}
    \caption{Determining the height of the point in reference to the trajectory center point for height-based autolabeling. 
    The figure illustrates points in one scan ring and the process is repeated at each scan ring. 
    Z is vertical and X is horizontal.} 
    \label{fig:height}
\end{figure}

The height-based autolabeling leverages the observation that the road is usually located lower than its surroundings, especially in winter driving conditions. 
The height $H_i$ for point $i$ is the difference between the point's vertical position and the trajectory center point vertical position $z_0$. 
If the point is below the reference, the height is set to zero (Fig. \ref{fig:height}). 

\begin{equation}
    H_i=
    \begin{cases}
    z_i-z_0, & \text{if } z_i>z_0 \\
    0, & \text{otherwise}
    \end{cases}
    \label{eq:height}
\end{equation}

For improved robustness, we reject points whose radial distance differs from the reference point by more than 5 m.   
To convert the height difference into a label in the range of 0-1, an exponential transform is applied.  

\begin{equation}
   l^{\text{h}}_i=\exp \left(-\frac{H_i^2}{\sigma_{\text{H}}^2}\right),
\end{equation}

where the parameter $\sigma_{\text{H}}$ controls the label's sensitivity to height variations. 
When the height difference between the road and its surroundings is high a larger value should be chosen and vice versa. 

\subsubsection{Gradient-based autolabeling}
\begin{figure}
    \centering
    \includegraphics[width=\linewidth]{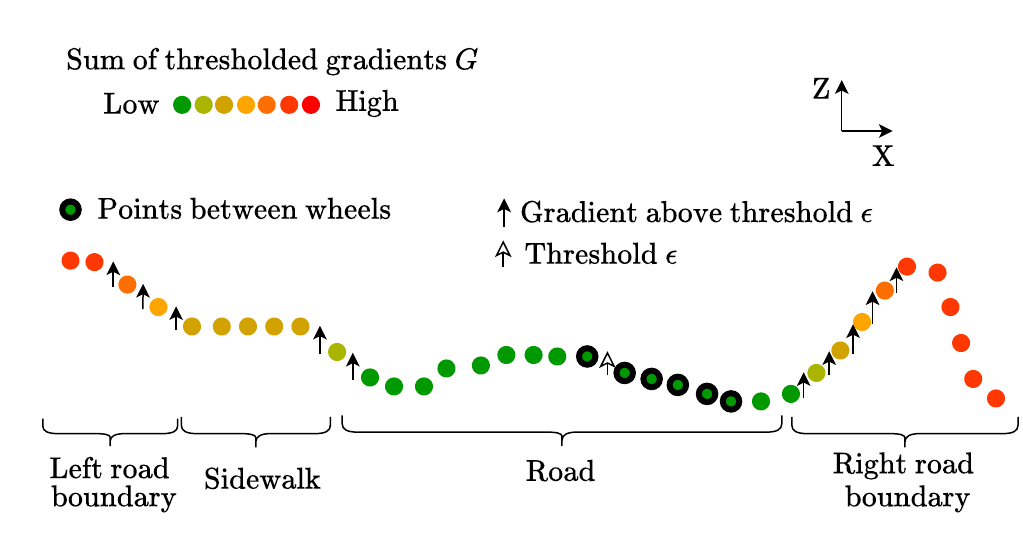}
    \caption{Determining the sum of thresholded gradients for gradient-based autolabeling. The figure illustrates points in one scan ring and the process is repeated at each scan ring. 
    Z is vertical and X is horizontal.}
    \label{fig:gradient}
\end{figure}

The gradient-based label is derived from the observation that road boundary areas usually have steep upward height changes causing large gradients, especially in winter driving conditions. 
Here the gradient is defined as the point-wise height difference $\Delta z$ between two consecutive points. 
Assuming movement along the scan ring from the center toward the edges of the road, the more upward gradients have already been encountered the less likely the point is road. 
To avoid false detections, small gradients caused by the road's uneven surface should be filtered out, but the unevenness varies, meaning a single threshold will not work for all cases. 
However, an adaptive estimate can be computed by finding the maximum gradient between the left and right wheel and setting it as the filtering threshold $\epsilon$. 
From these premises, we propose the sum of thresholded gradients $G_i$ for estimating if a given point belongs to the road. 
For point $i$ it is defined as the sum of gradients $\Delta z$ between the trajectory center and the point $i$ itself that are above the threshold $\epsilon$ (Fig. \ref{fig:gradient}).  

\begin{equation}
\begin{aligned}
 G_i & = \sum_{j=0}^{i} g_j, \quad \text{where} \\
 g_j & = 
\begin{cases} 
\Delta z_j, & \text{if } \Delta z_j \geq \epsilon \\
0 & \text{otherwise}
\end{cases}
\end{aligned}
\end{equation}

To convert the sum of thresholded gradients into a label in the range of 0-1 an exponential transform is applied.

\begin{equation}
   l^{\text{grad}}_i=\exp \left(-\frac{G_i^2}{\sigma_{\text{G}}^2}\right),
\end{equation}

where the parameter $\sigma_{\text{G}}$ controls the label's sensitivity to gradient variations. 
When the road is separated from the background by clear boundaries larger values should be used and vice versa. 

\subsection{Camera-based autolabeling}
\label{sec:camera}
The camera-based autolabeling leverages the fact that road areas have different visual appearance compared to the background. 
Here the visual appearance is evaluated using the pre-trained vision foundation model DinoV2-g. 
Image input size is 1224x400 pixels and each 14x14 pixel patch is assigned a feature that describes the contents of that patch. 
The road prototype $\boldsymbol{f}_{\text{proto}}$ is defined as the mean of all features that belong to the trajectory. 
It is computed based on the current frame, except when less than 200 trajectory patches are available. 
In that case, the prototype from the last frame with at least 200 trajectory patches is used. 
The similarity between each patch $\boldsymbol{f}_i$ and the prototype $\boldsymbol{f}_{\text{proto}}$ is defined based on cosine similarity.   

\begin{equation}
\label{attention}
\begin{aligned}
C_{i}=\frac{\boldsymbol{f}_{i} \cdot \boldsymbol{f}_{\text{proto}}}{\|\boldsymbol{f}_{i}\| \: \|\boldsymbol{f}_{\text{proto}}\|},
\end{aligned}
\end{equation}

The cosine similarity $C_i$ is normalized by dividing with the maximum similarity inside the frame. 
The label value for each patch is computed from the normalized similarity $C_i^{\text{norm}}$ with an exponential transform:

\begin{equation}
   l^{\text{cam}}_i=\exp\left(-\frac{(1-C_{i}^{\text{norm}})^2}{\sigma_{\text{C}}^2}\right)
\end{equation}

where the parameter $\sigma_{\text{C}}$ controls the label's sensitivity to changes in visual appearance. 
When the road appearance has high variation a larger value should be used and vice versa. 
The patch-wise label is interpolated to the original image resolution to yield a pixel-wise label.  

\subsection{Data}

Our method was validated using a dataset collected in Finland in February 2024. Data acquisition took place across two distinct environments: suburban and countryside roads. 
The data was gathered with a research vehicle equipped with Novatel PWRPAK 7DE2 GNSS/INS unit, FLIR blackfly S 2448x2048 camera and Velodyne VLP-32C top mounted 3D-lidar. 
Intrinsic parameters and radial distortion coefficients were calibrated with MATLAB's checkerboard calibration tool. 
The extrinsic translation between the lidar and camera frames was measured, and rotations were manually refined to optimize alignment in the validation set. 
Sensor readings were synchronized based on timestamps and frames where windshield wipers were visible were removed. The trajectory was determined based on the GNSS/INS unit estimates. 

For training and evaluation, data sequences from each environment were split into train, validation, and test sets, with unique road segments in each split. 
Balanced data distribution was achieved by sampling every 5 meters in the train set and every 10 meters in the validation and test sets. 
Each scene included 100 validation and 400 test images, all manually labeled. 
The training set consisted of approximately 7k unlabeled samples for the suburban scene and 6k unlabeled samples for the countryside scene. The labeled test and validation samples are made publicly available. 

\subsection{Validation}
We evaluate our method's road segmentation performance against recent baselines on our winter driving dataset. 
All image-based methods use an input size of 1224x400 and for each method we choose parameters that yield the highest Intersection over Union (IoU) on the validation set. 
The rest of the implementation details are presented below.

\textbf{OURS} is trained solely with our autolabels generated for the whole train set with parameters: $\sigma_{\text{C}}=0.6$, $\sigma_{\text{H}}=0.1$, $\sigma_{\text{G}}=0.02$. 
The prediction model is Deeplabv3 \cite{chen2017rethinking} with a ResNet50 backbone, trained with Adam optimizer (batch size 32, 60 epochs, learning rate 1e-3). 
\textbf{Supervised} is trained on 200 manually labeled validation images via cross-validation, using the same hyperparameters and model as OURS to demonstrate performance achievable by a limited number of manual labels in contrast to a high number of autolabels.    
\textbf{SAM2\cite{ravi2024sam}} generates a set of segmentation mask proposals around a given input point, each associated with a quality score. We prompt SAM2 with a fixed point taken in front of the vehicle (50 pixels from the bottom, center of the image). In our driving scenarios, this point always lies inside the road. From the generated proposals, we choose the mask with the highest quality score.  
\textbf{Seo et. al.\cite{seo2023learning}} is trained with the whole train set. 
The learning rate is set to 1e-4 and the batch size to 16, with other parameters as in the original publication. 
\textbf{Lidar Boundary Detection\cite{wang2020speed}} is configured to match our hardware specifications with 32 scan rings and 0.2-degree angular resolution. 
The ground point segmentation distance threshold was set to 2 m and the scan was limited to a 90-degree field of view as with our autolabels. 
Using the full scan reduced performance. 
Other parameters follow the original publication.

\section{Results}

\begin{figure*}[]
\centering
\begin{tabular}{@{}c@{} c@{} c@{} c@{} c@{} c@{}}

\bf{Trajectory} &
\bf{Camera-based (C)} &
\bf{Height-based (H)} &
\bf{Gradient-based (G)} &
\bf{H + G} &
\bf{C + H + G + CRF} \\

\includegraphics[width = 1.2in]{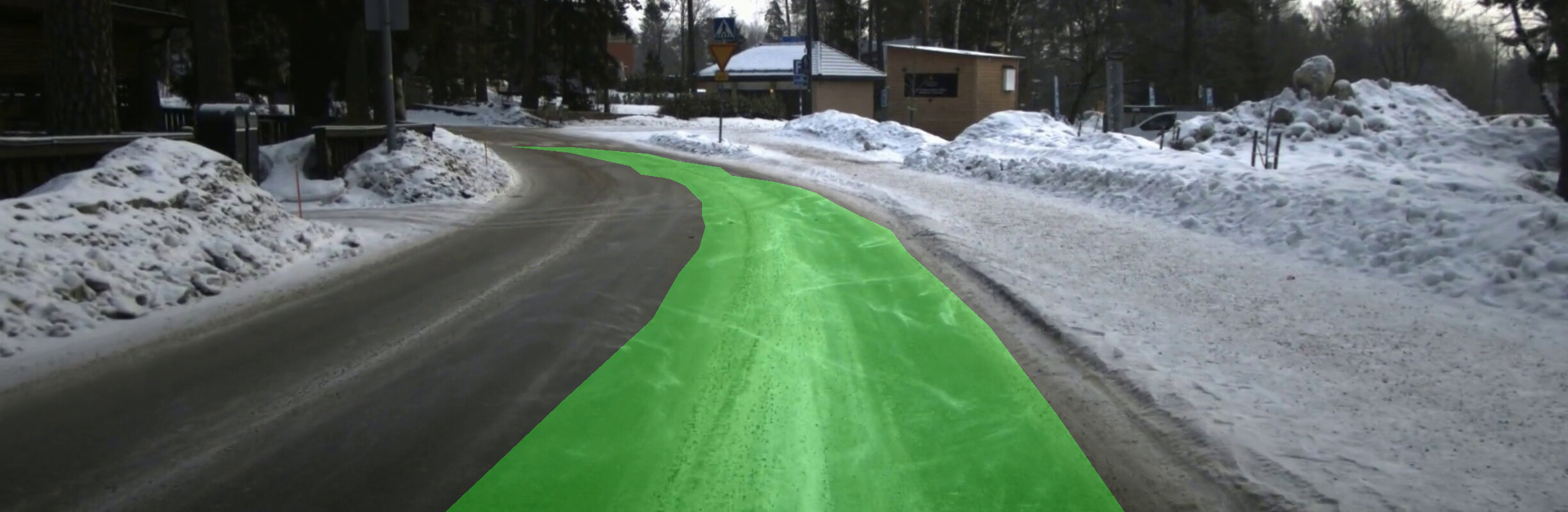} &
\includegraphics[width = 1.2in]{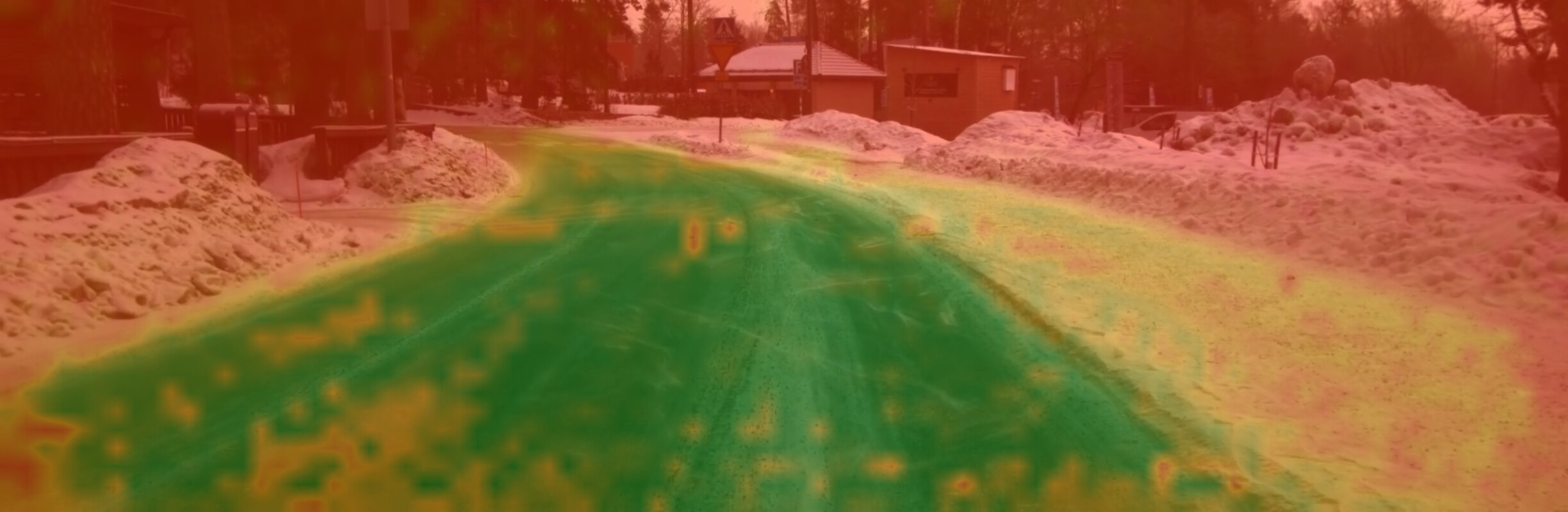} &
\includegraphics[width = 1.2in]{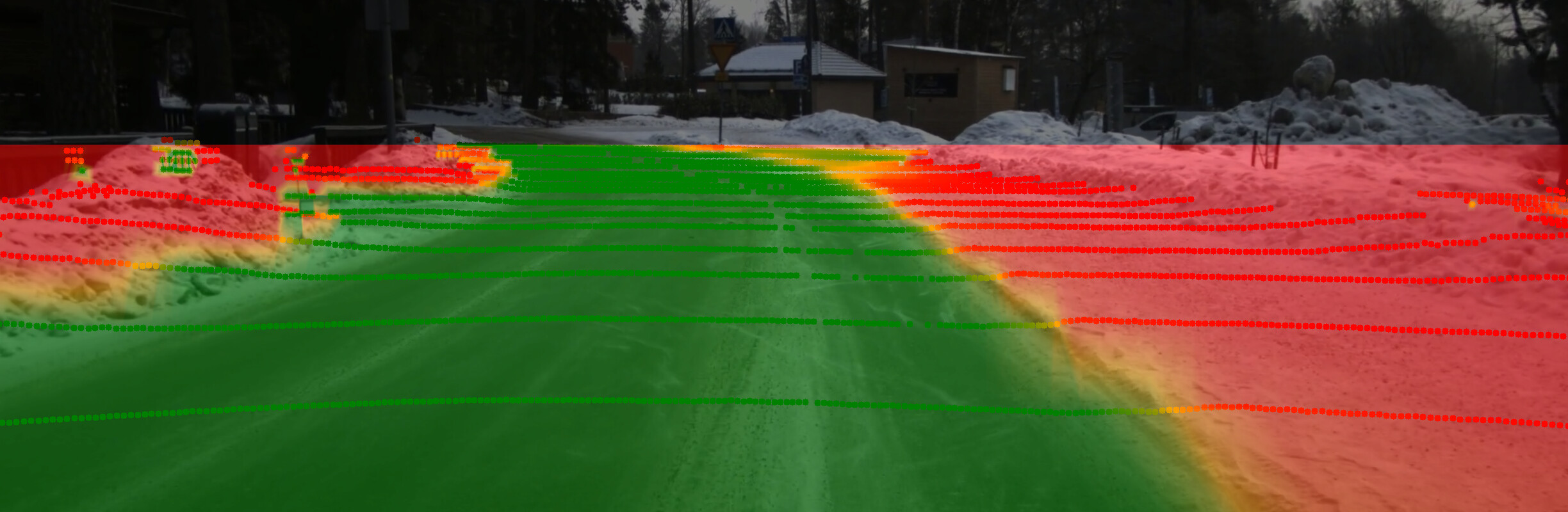} &
\includegraphics[width = 1.2in]{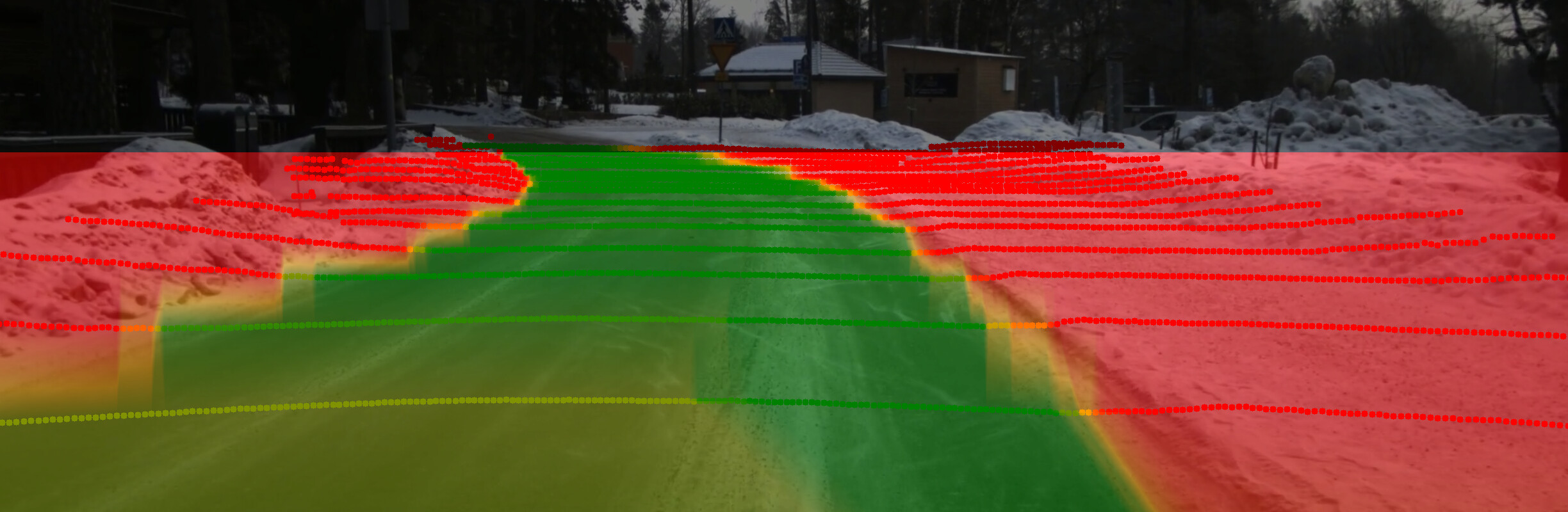} &
\includegraphics[width = 1.2in]{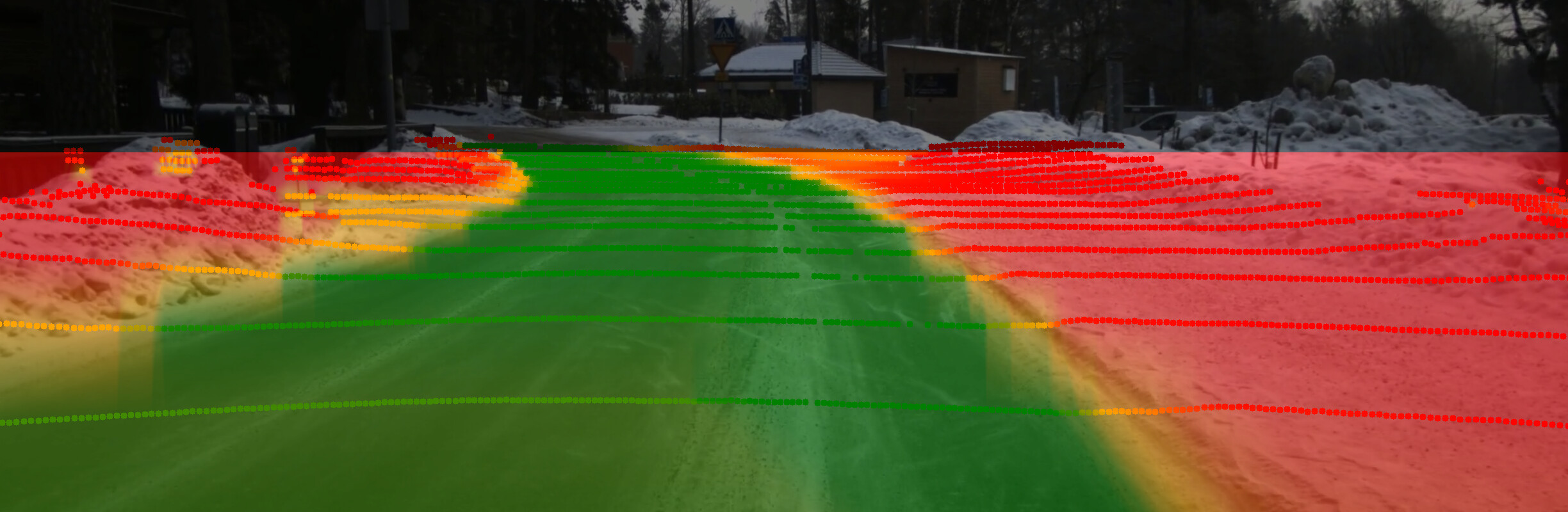} &
\includegraphics[width = 1.2in]{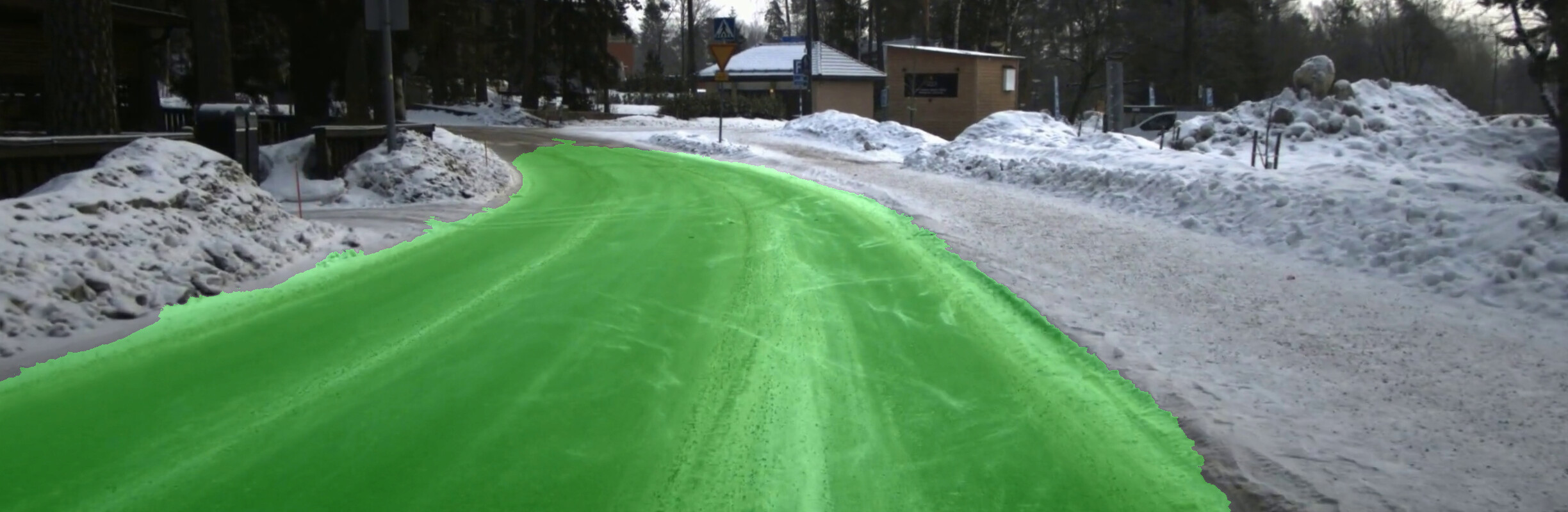} \\

\includegraphics[width = 1.2in]{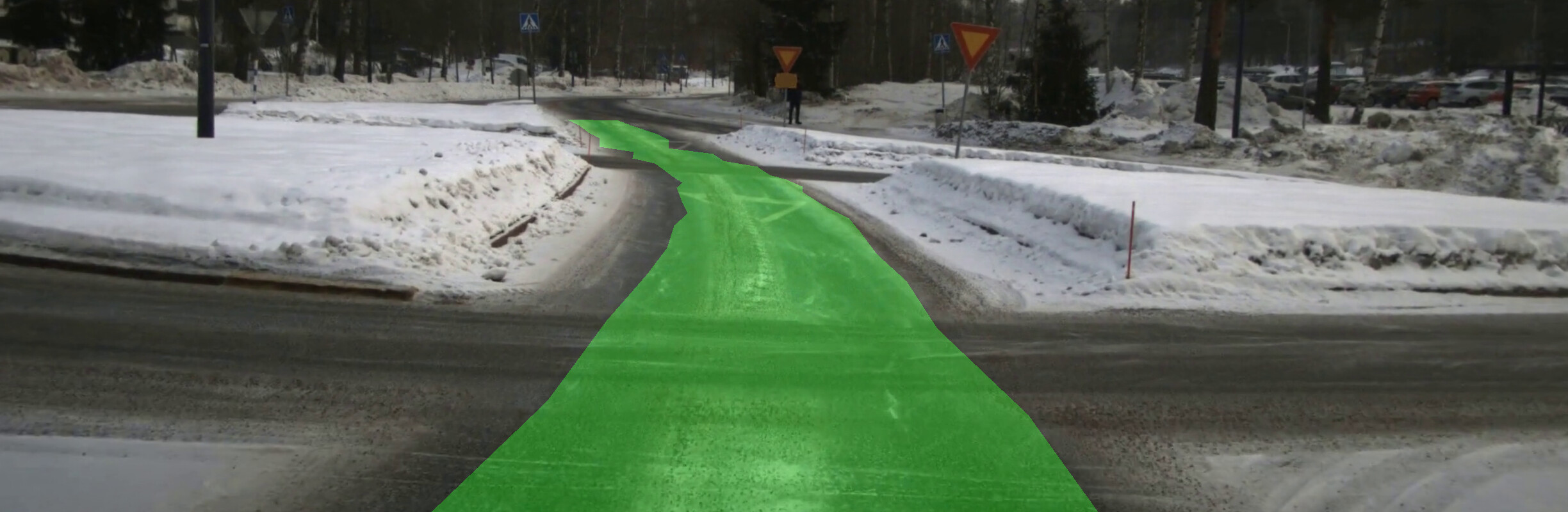} &
\includegraphics[width = 1.2in]{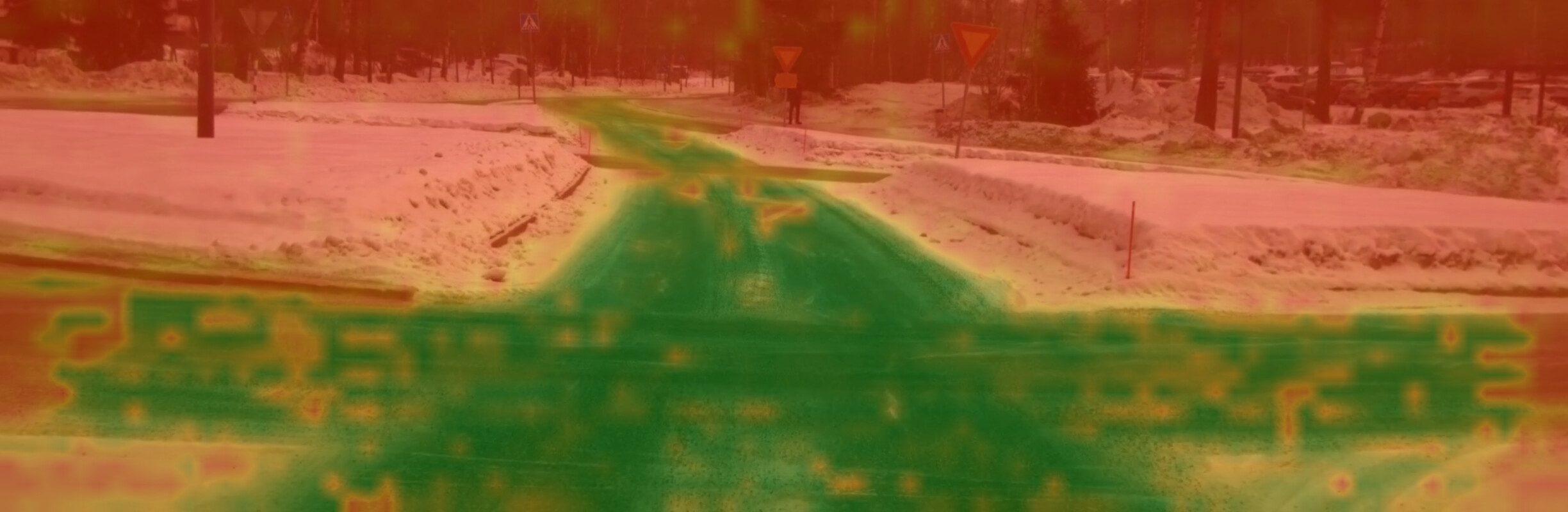} &
\includegraphics[width = 1.2in]{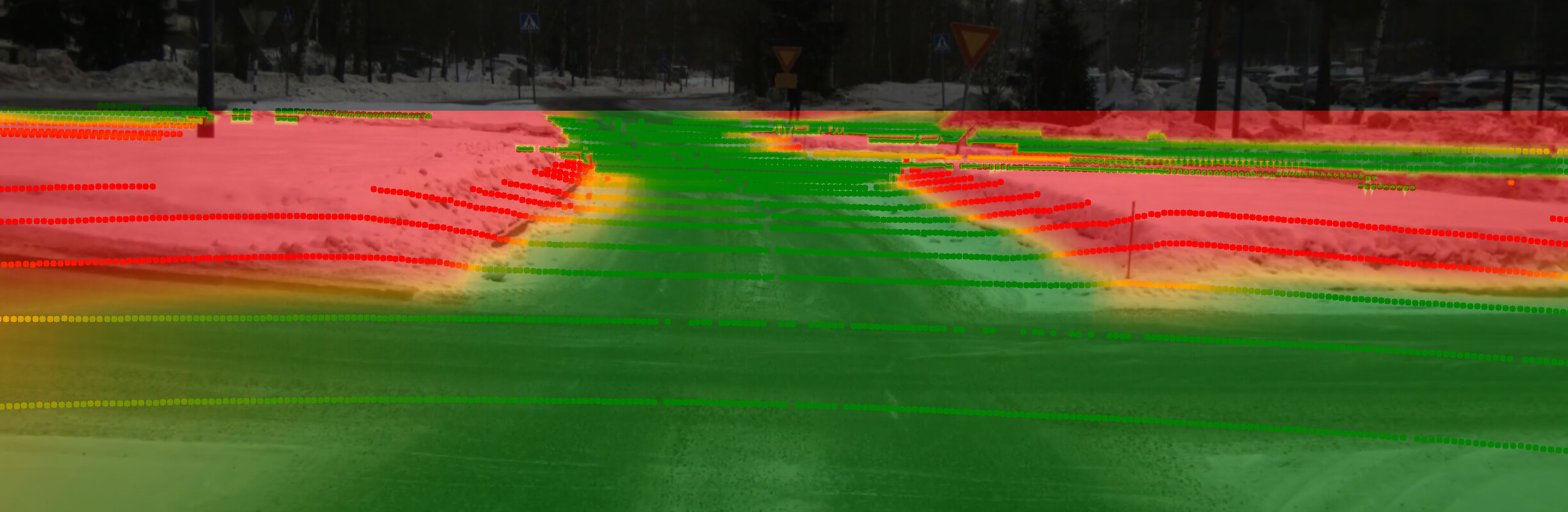} &
\includegraphics[width = 1.2in]{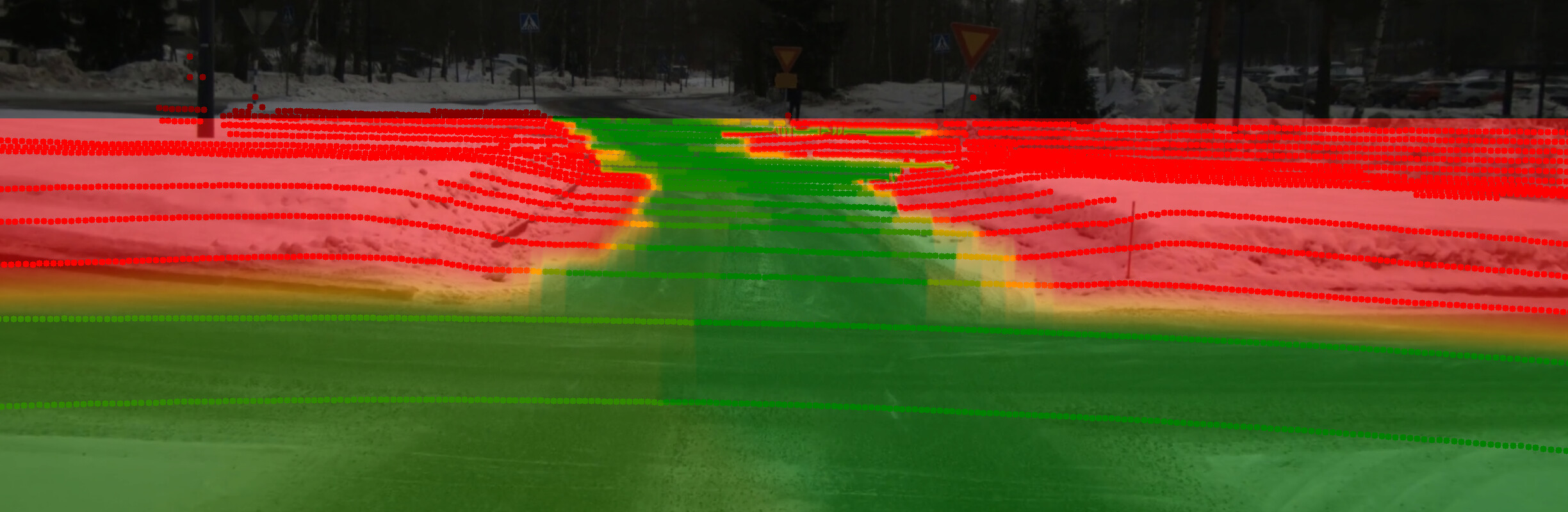} &
\includegraphics[width = 1.2in]{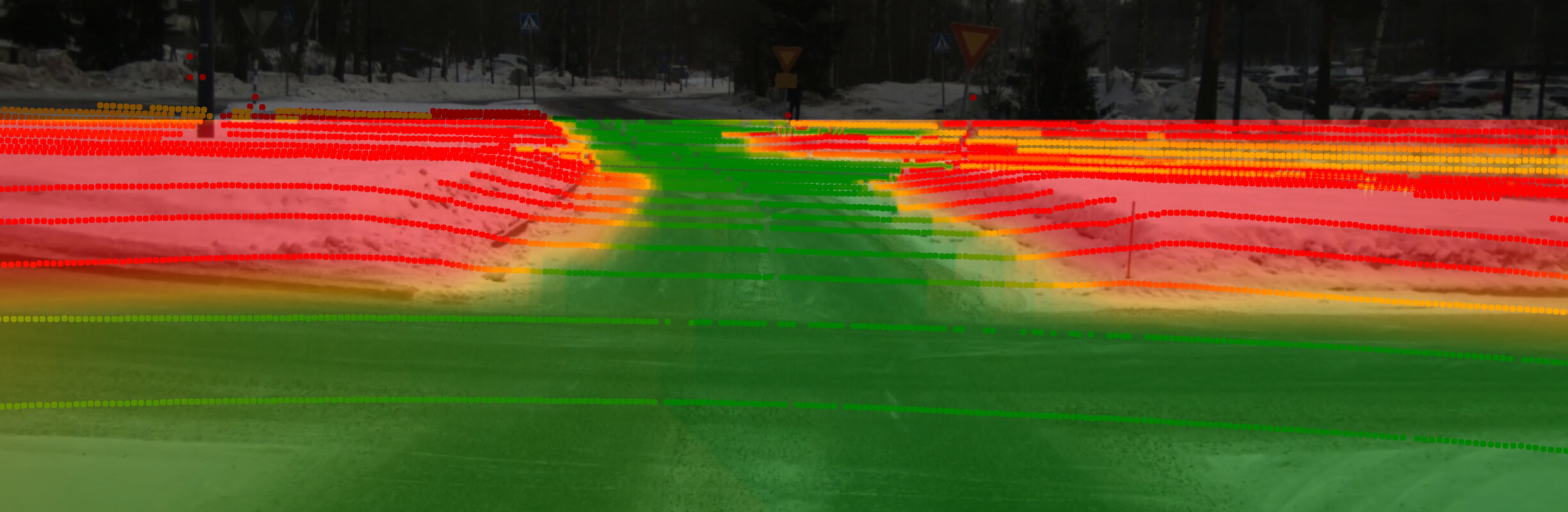} &
\includegraphics[width = 1.2in]{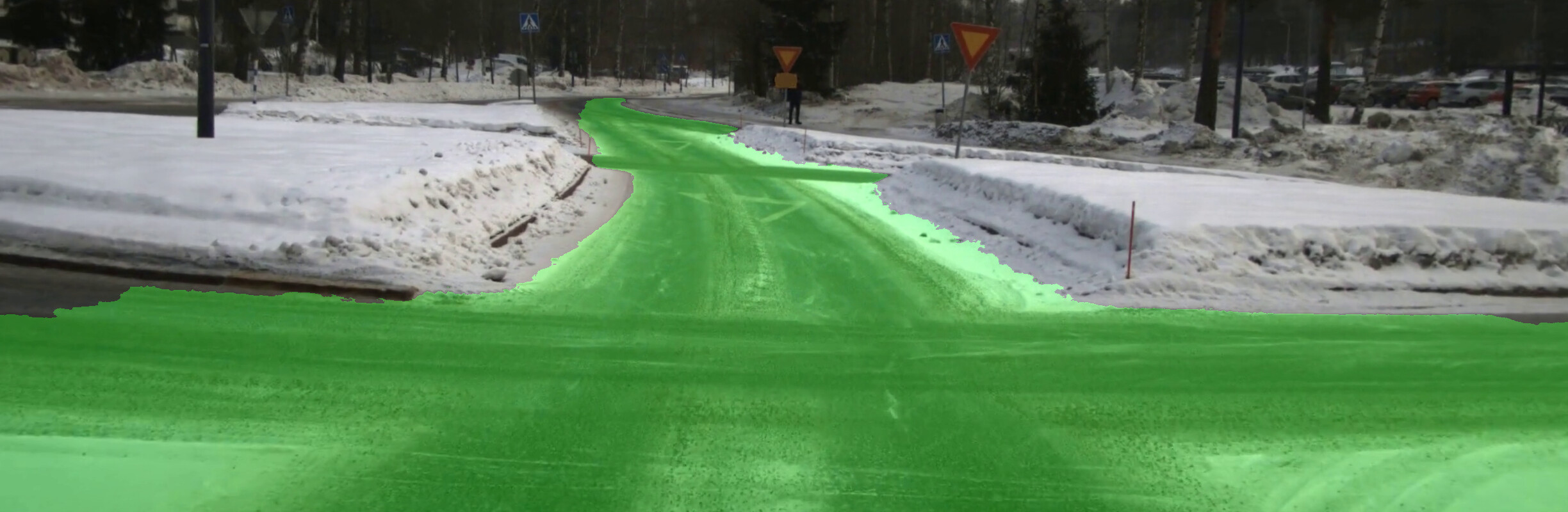} \\

\includegraphics[width = 1.2in]{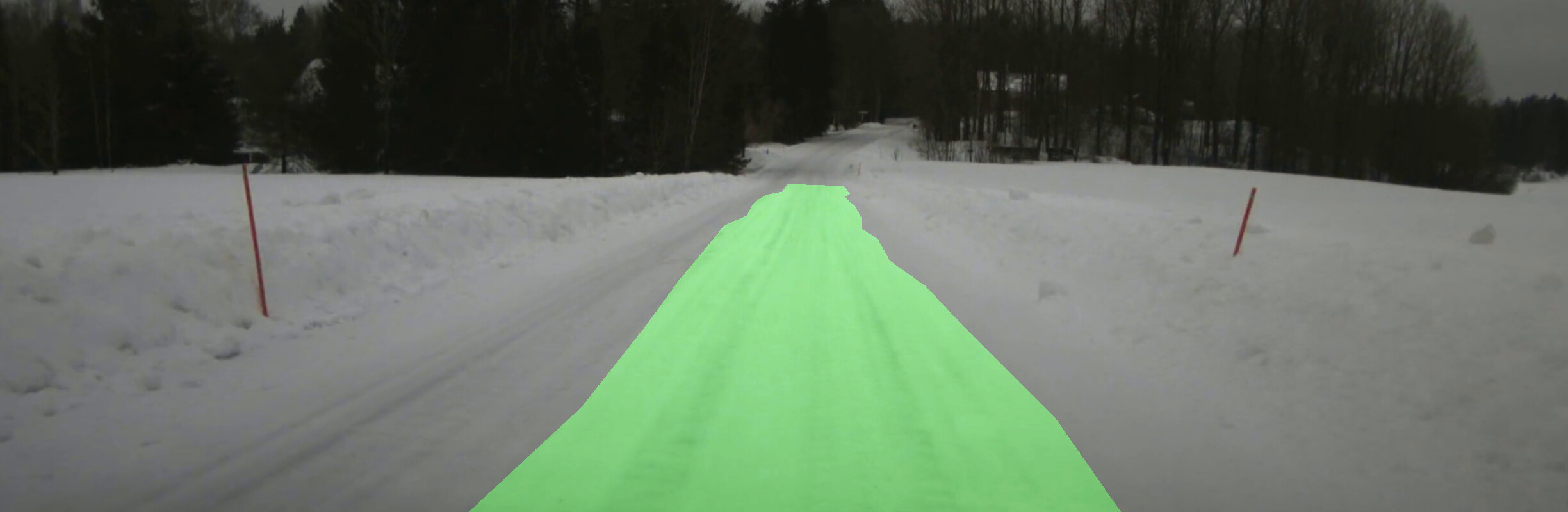} &
\includegraphics[width = 1.2in]{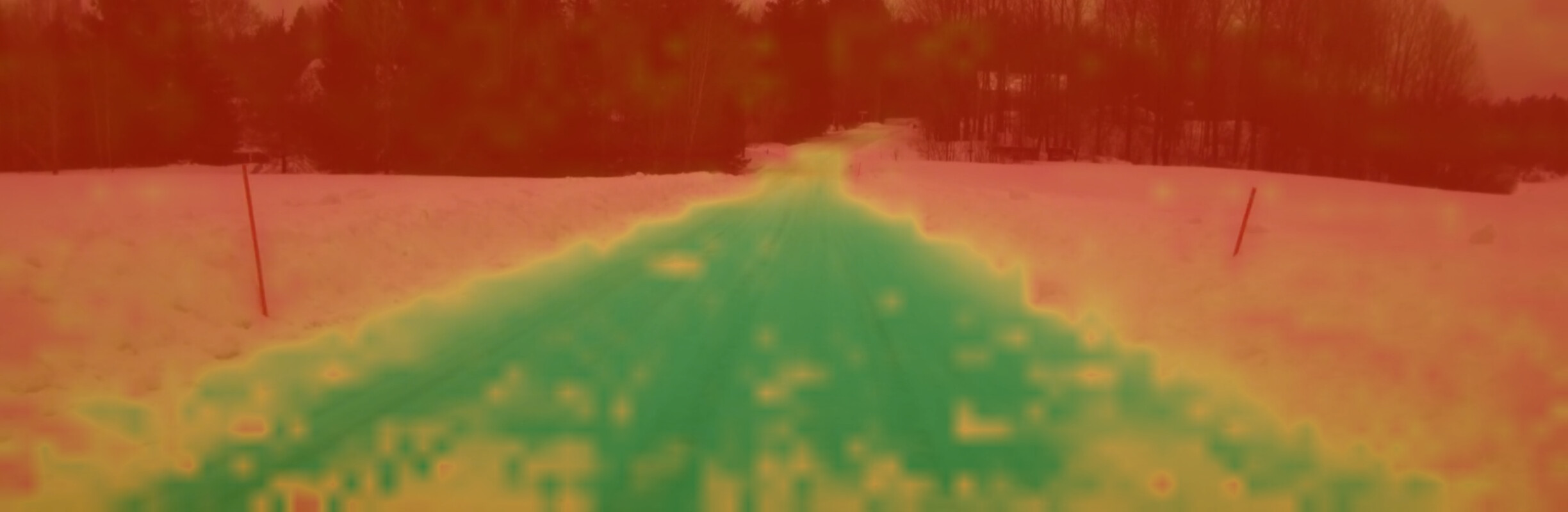} &
\includegraphics[width = 1.2in]{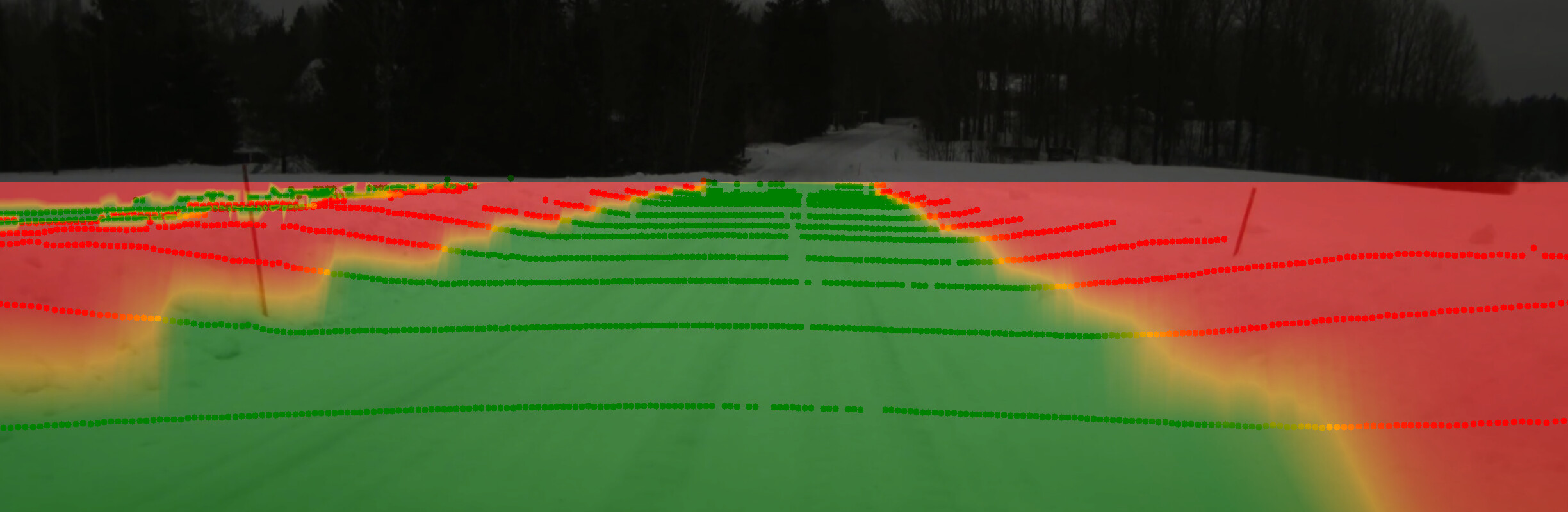} &
\includegraphics[width = 1.2in]{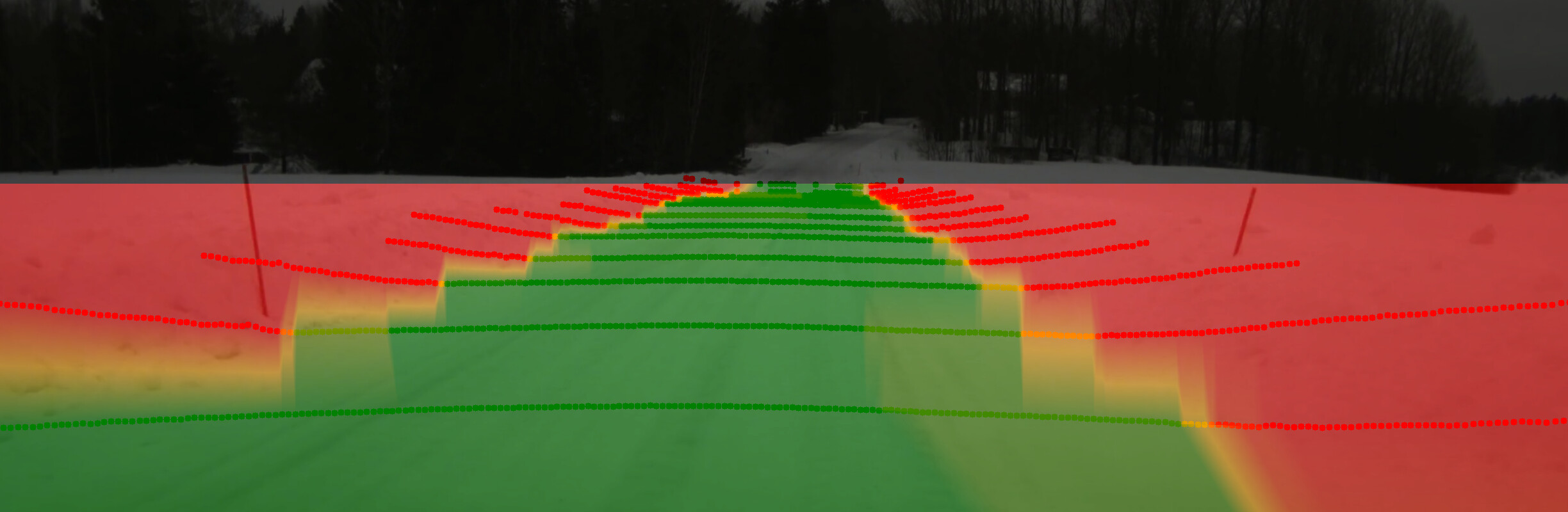} &
\includegraphics[width = 1.2in]{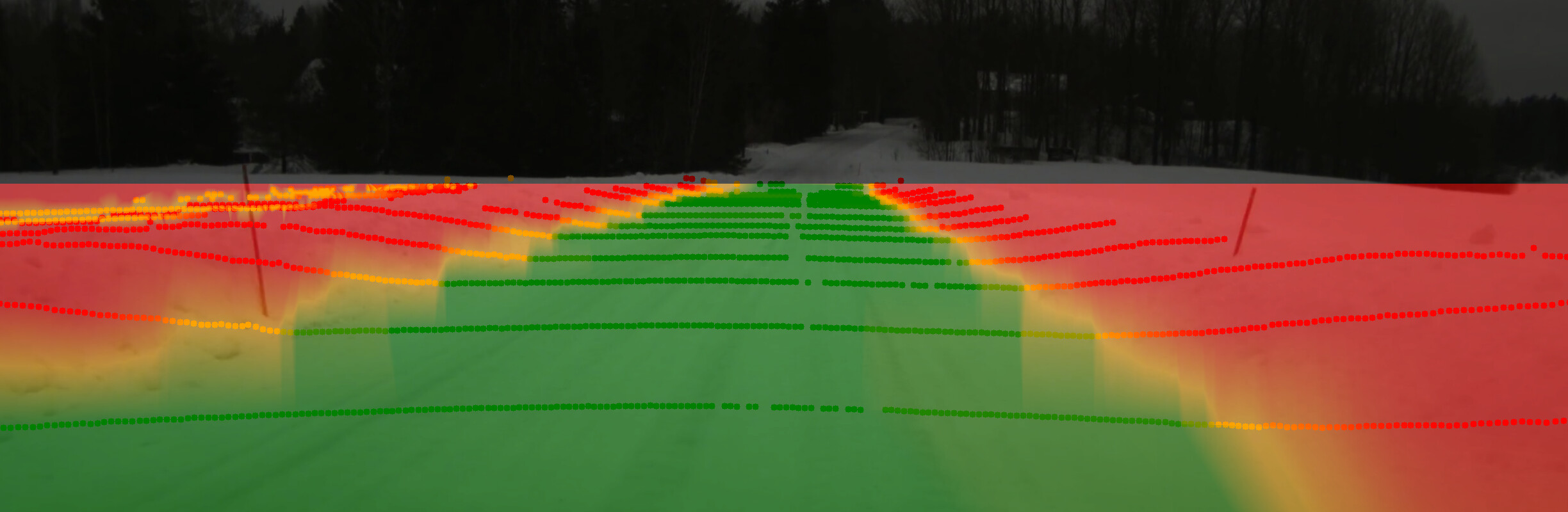} &
\includegraphics[width = 1.2in]{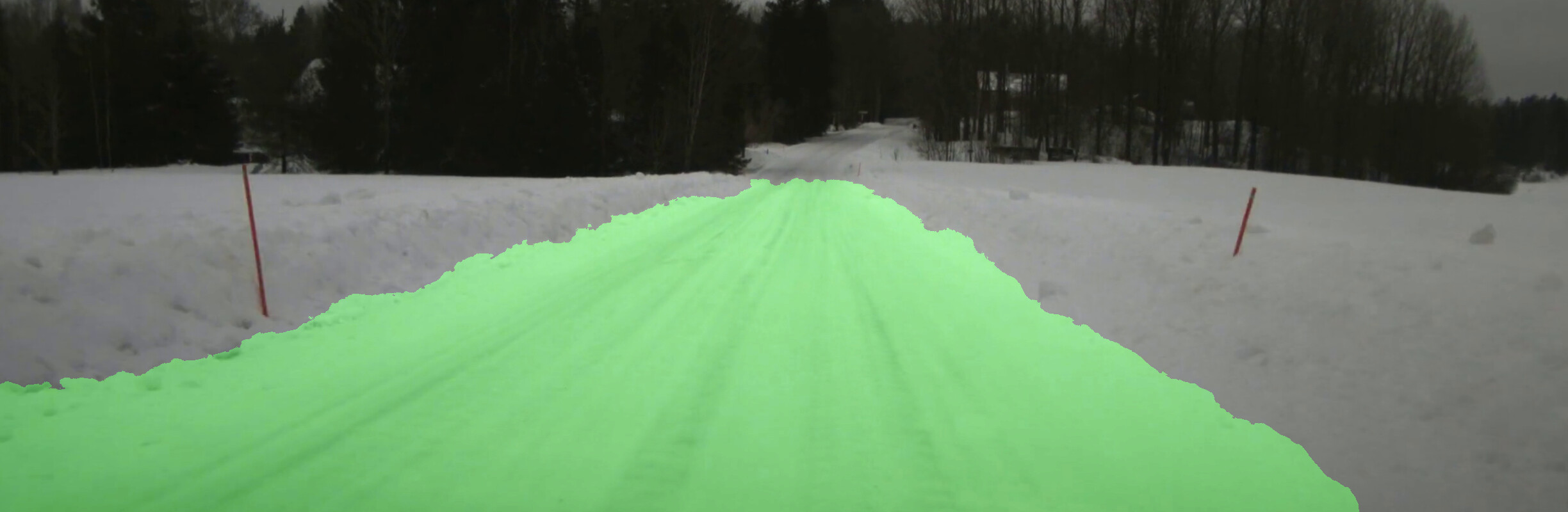} \\

\includegraphics[width = 1.2in]{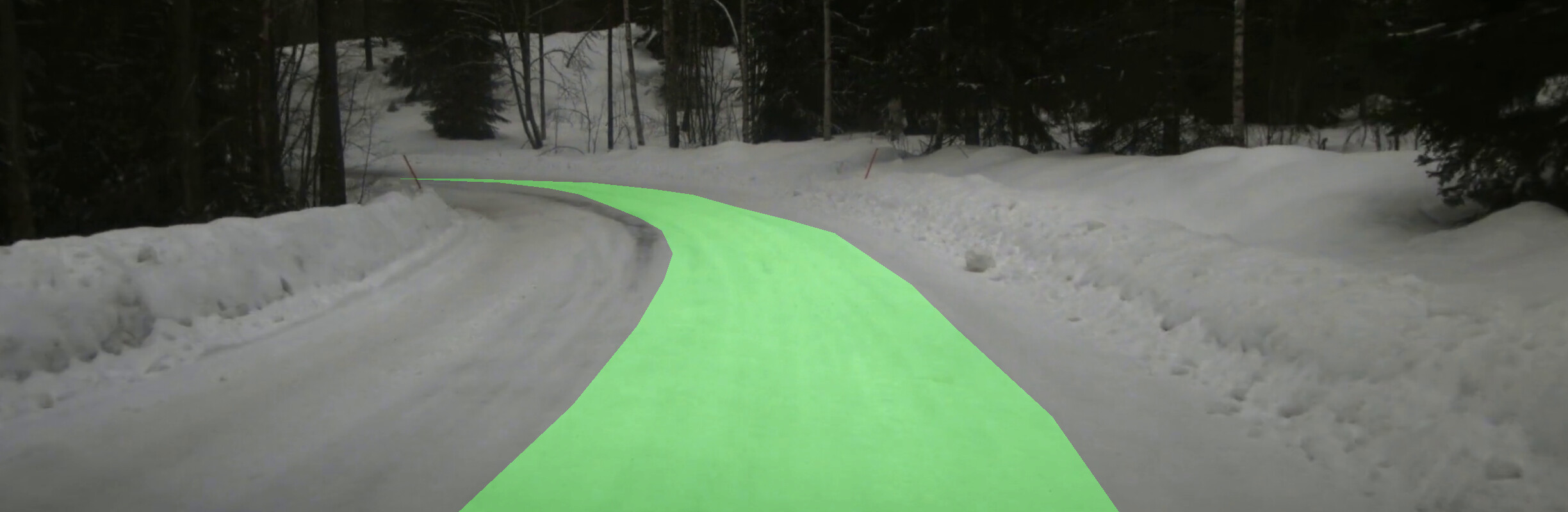} &
\includegraphics[width = 1.2in]{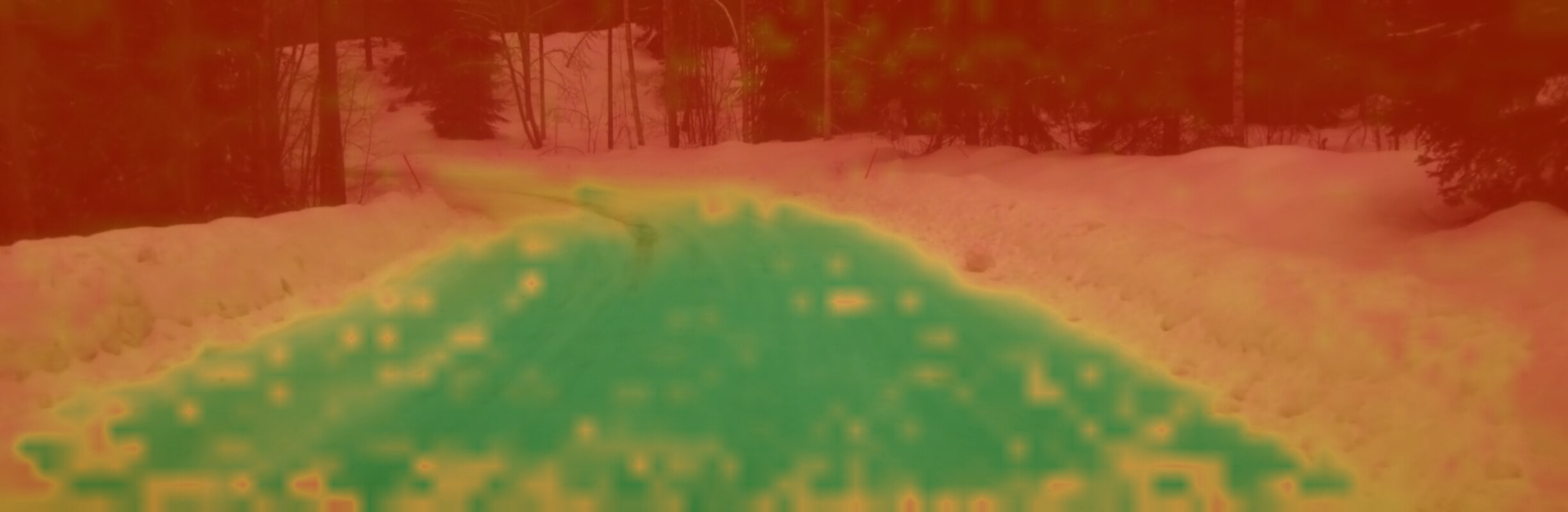} &
\includegraphics[width = 1.2in]{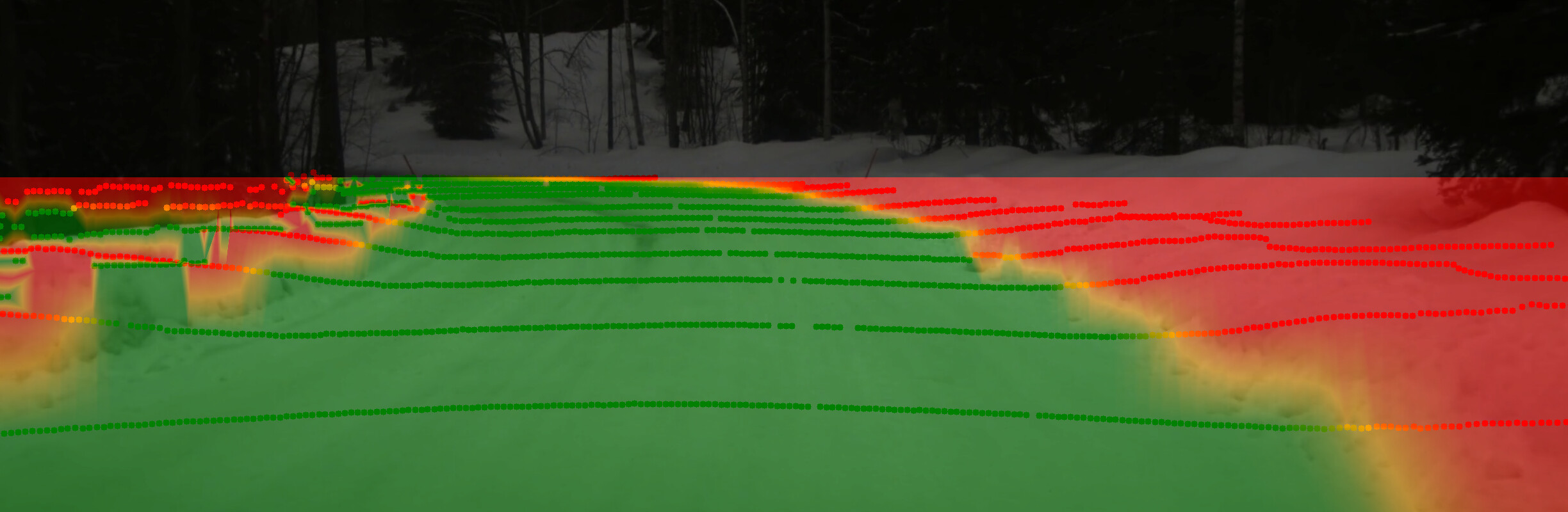} &
\includegraphics[width = 1.2in]{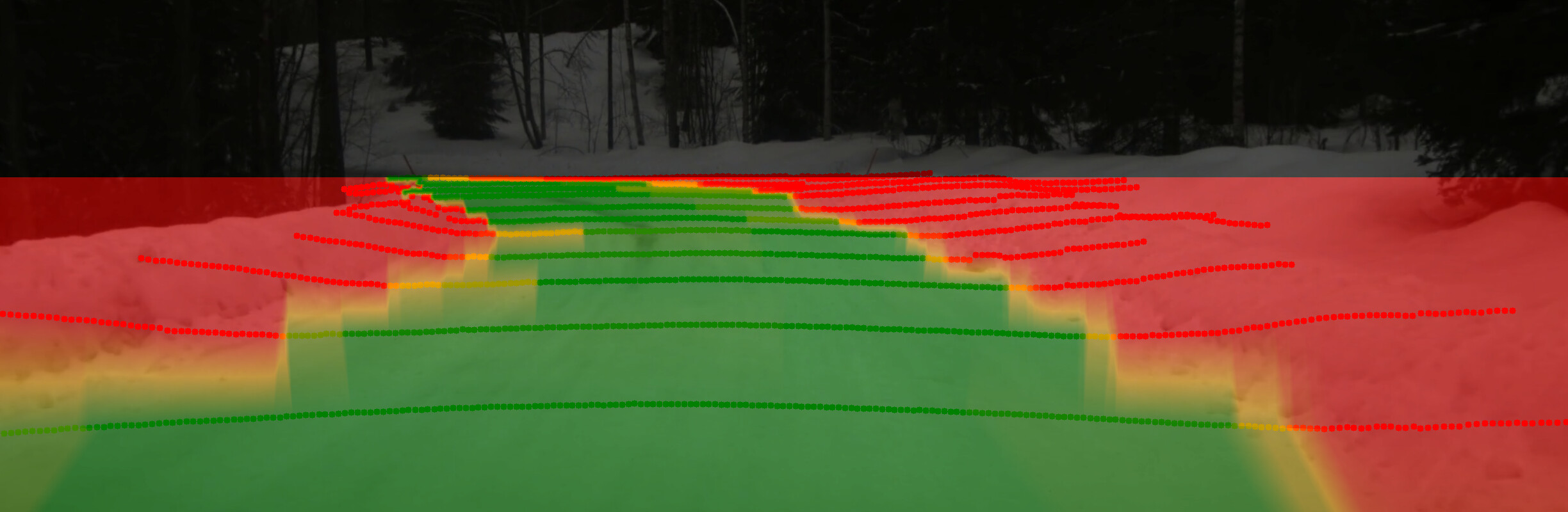} &
\includegraphics[width = 1.2in]{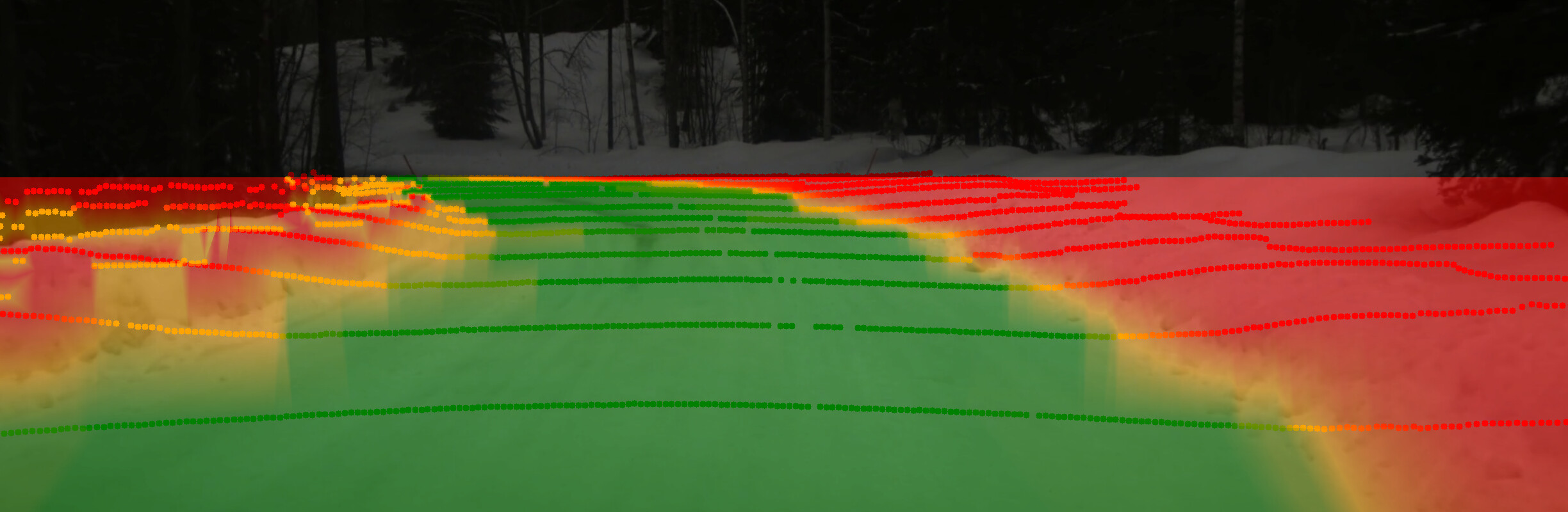} &
\includegraphics[width = 1.2in]{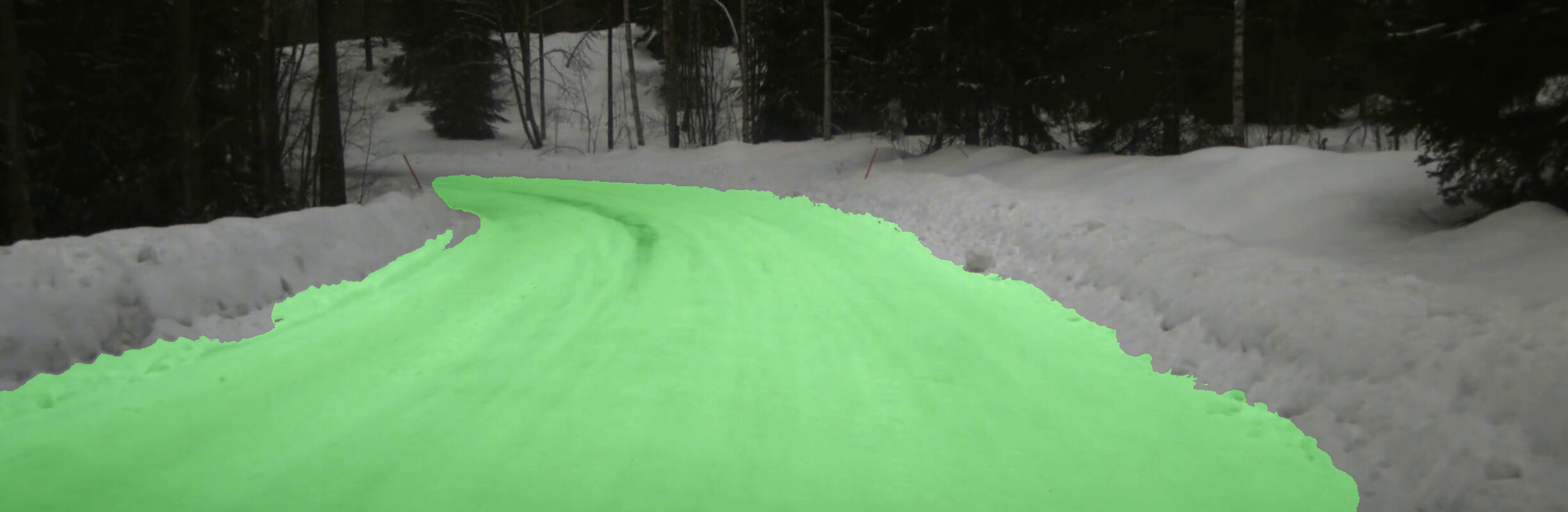} \\

\end{tabular}
\caption{Autolabels generated with different configurations.}
\label{fig:labels}
\end{figure*}

We use IoU to evaluate the overall quality of the road predictions. Precision (PRE), Recall (REC), and F1-score (F1) are also reported. The efficacy of each proposed autolabeling component was evaluated through an ablation study (Table \ref{tab:ablations}). We also present example labels generated with different configurations (Fig. \ref{fig:labels}). 
Using only the camera-based label resulted in an IoU of 83.5. 
Adding the gradient-based label increased IoU by +0.8, while including the height-based label provided an additional +6.5 IoU. 
Incorporating CRF further improved IoU by +0.4, bringing the final autolabeling accuracy to 90.2 IoU.

\begin{figure*}[]
\centering
\begin{tabular}{@{}c@{} c@{} c@{} c@{} c@{}}

&
&
\bf{Lidar Boundary} &
&
\\

\bf{Seo et. al.} &
\bf{SAM2} &
\bf{Detection} &
\bf{Supervised} &
\bf{OURS} \\

\includegraphics[width = 1.4in]{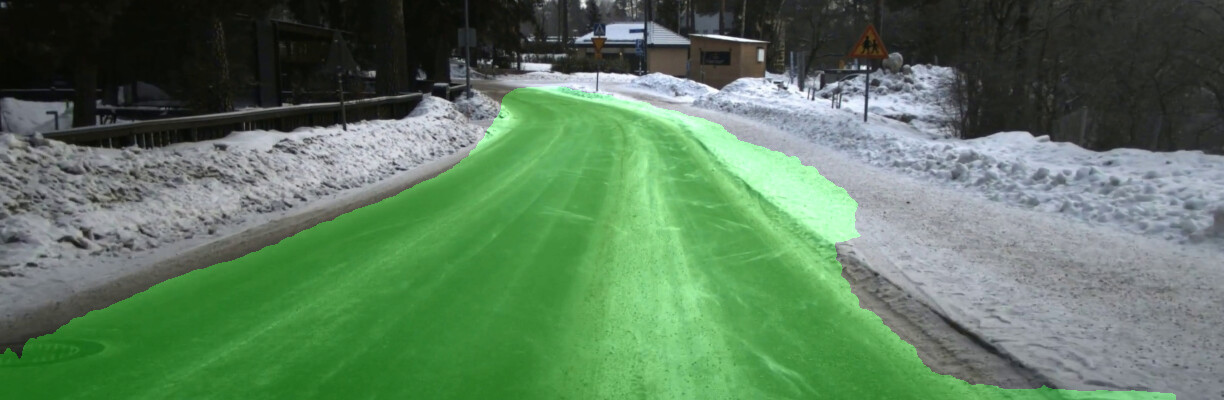} &
\includegraphics[width = 1.4in]{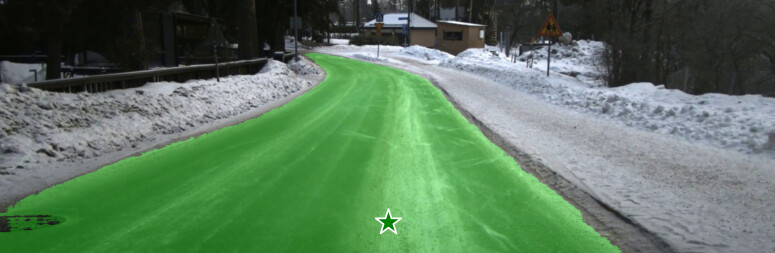} &
\includegraphics[width = 1.4in]{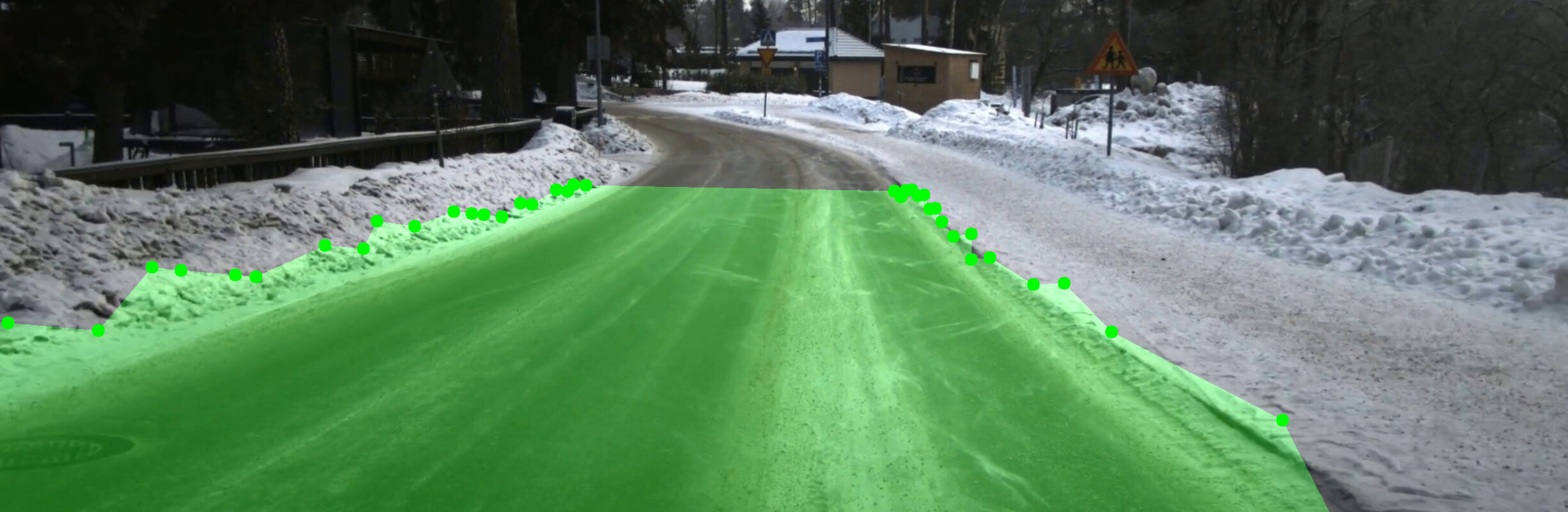} &
\includegraphics[width = 1.4in]{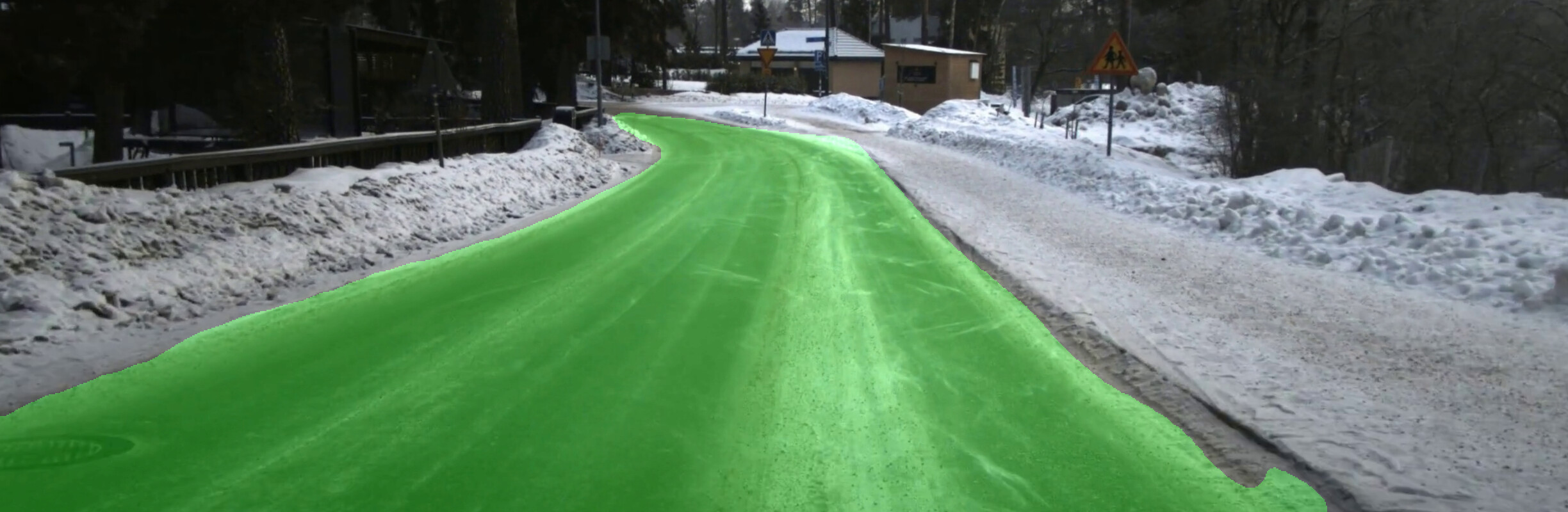} &
\includegraphics[width = 1.4in]{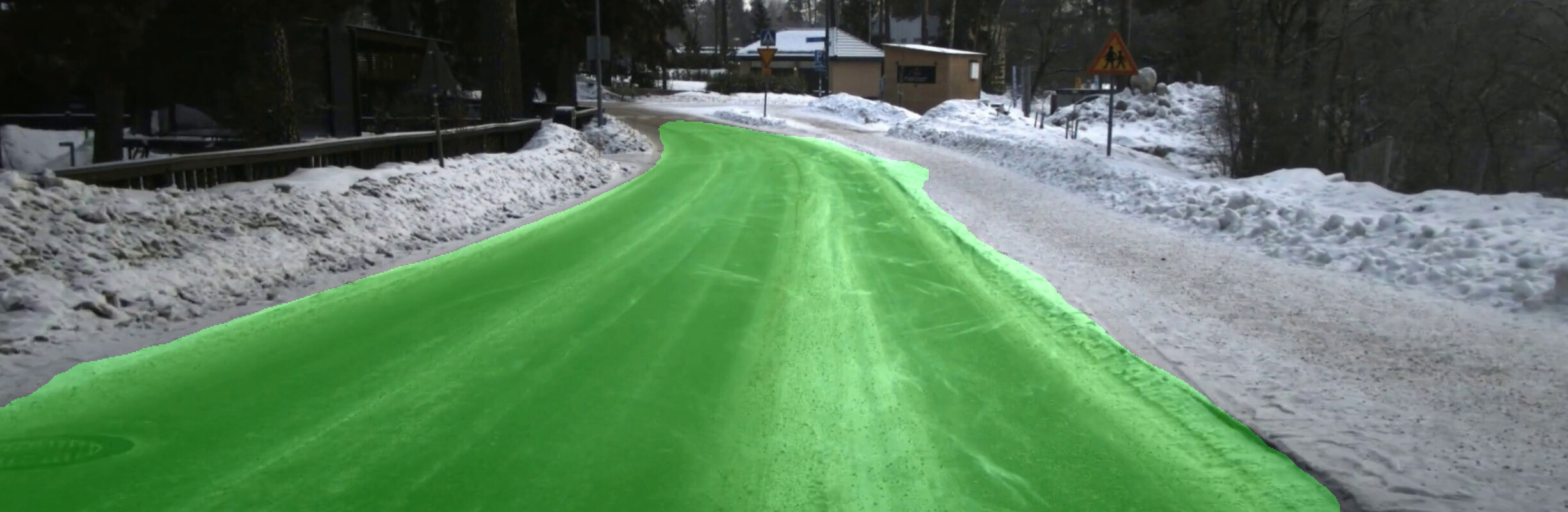} \\

\includegraphics[width = 1.4in]{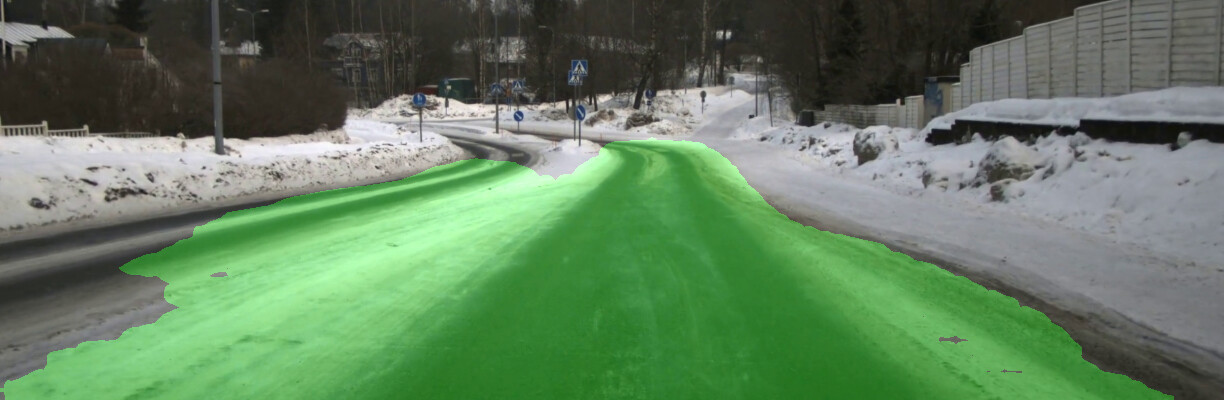} &
\includegraphics[width = 1.4in]{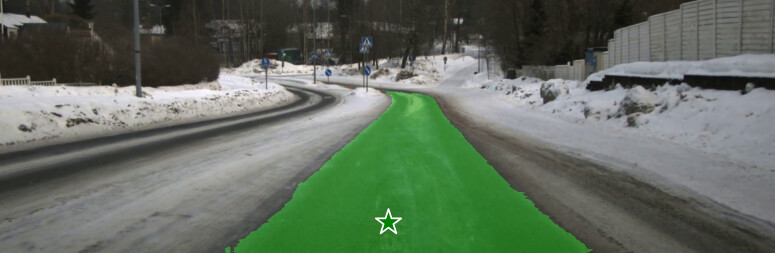} &
\includegraphics[width = 1.4in]{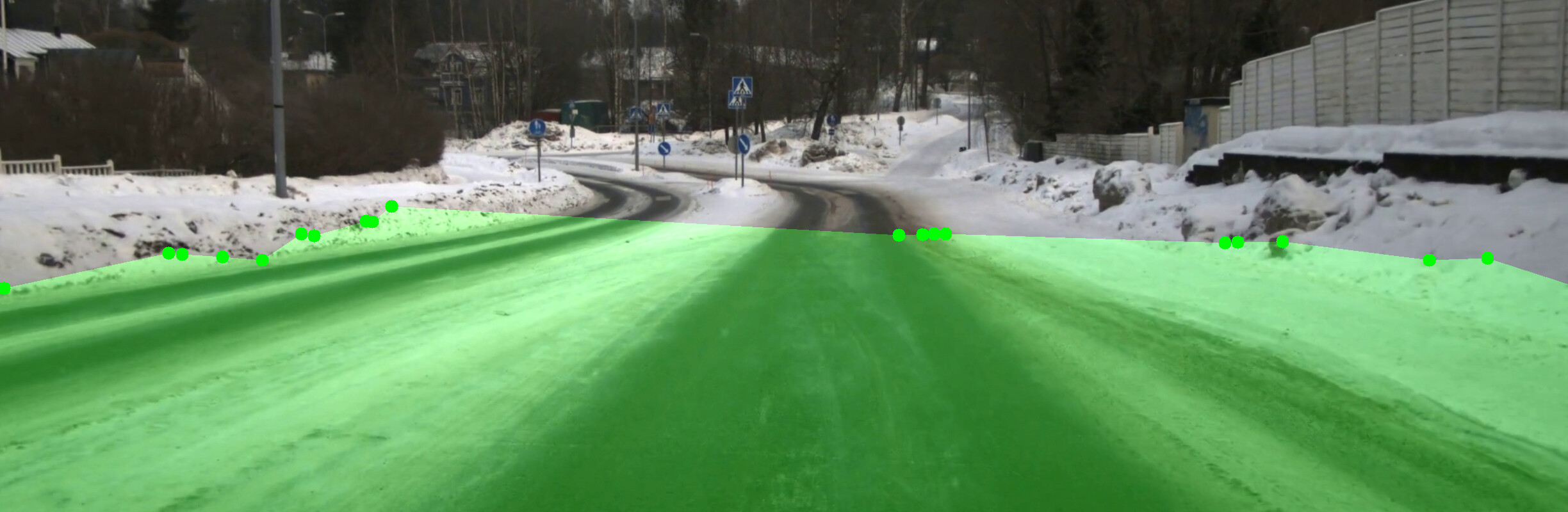} &
\includegraphics[width = 1.4in]{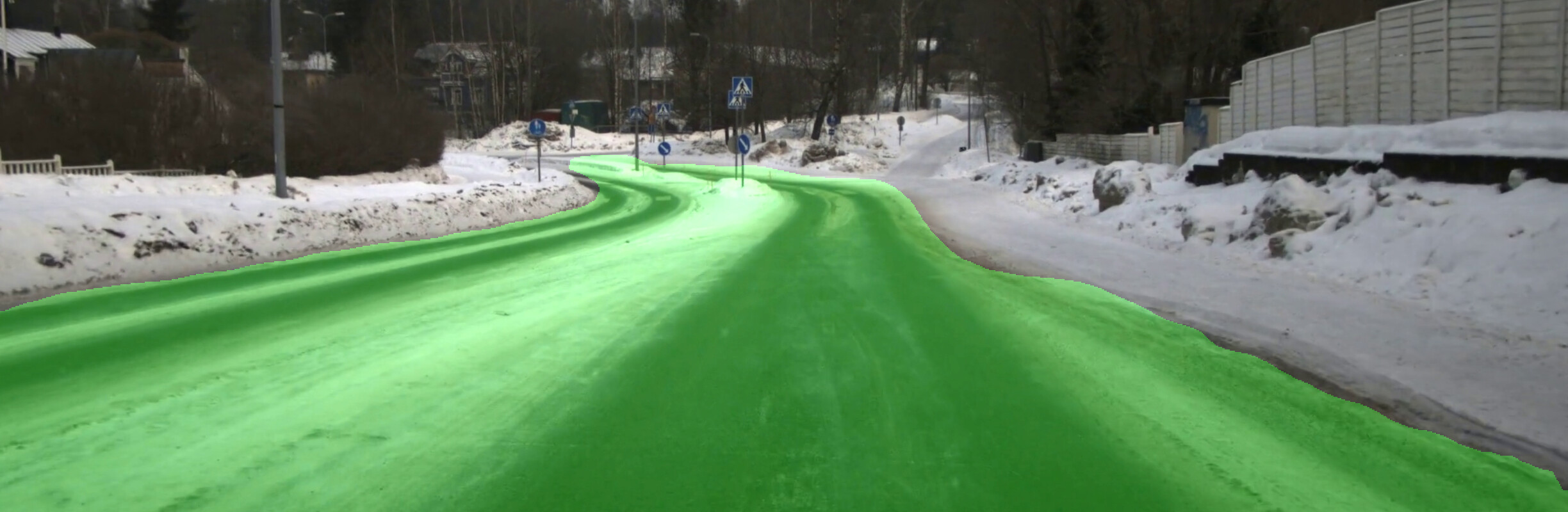} &
\includegraphics[width = 1.4in]{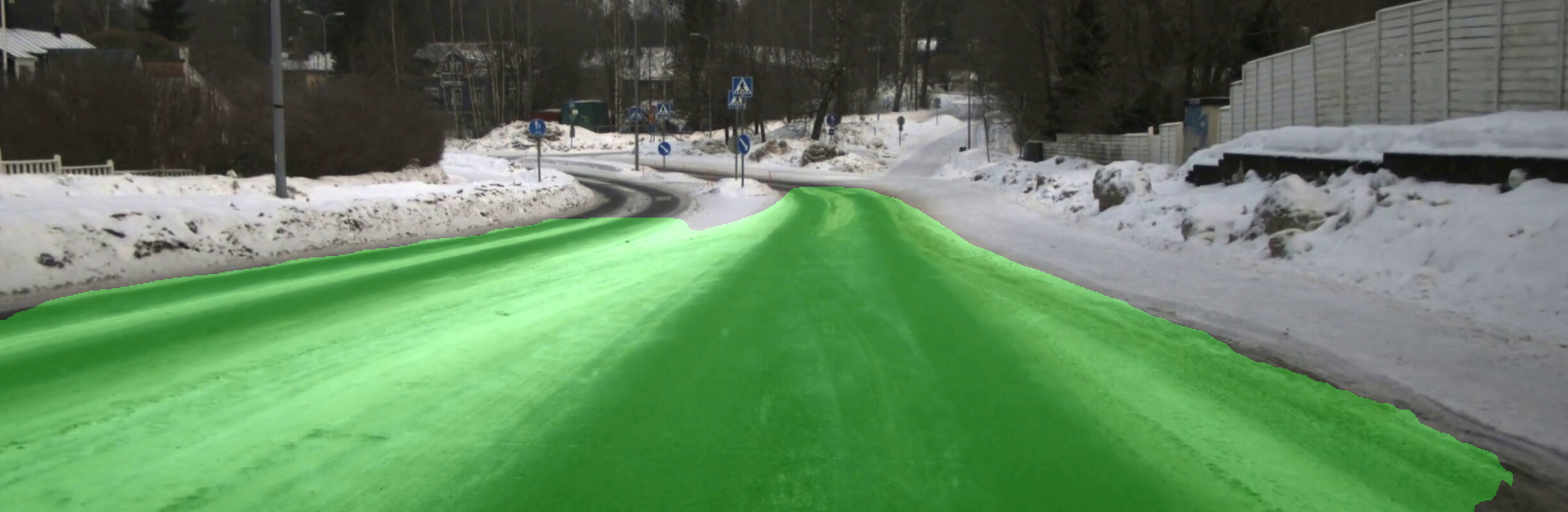} \\

\includegraphics[width = 1.4in]{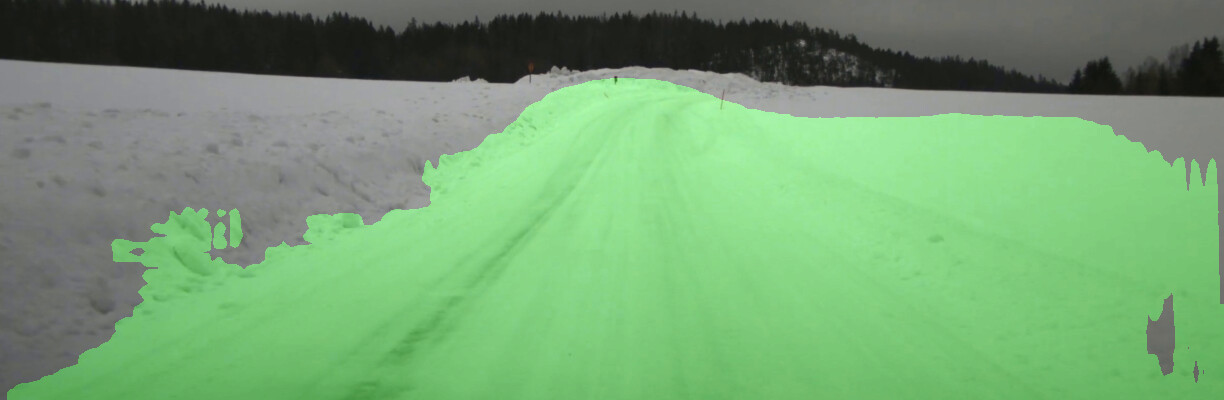} &
\includegraphics[width = 1.4in]{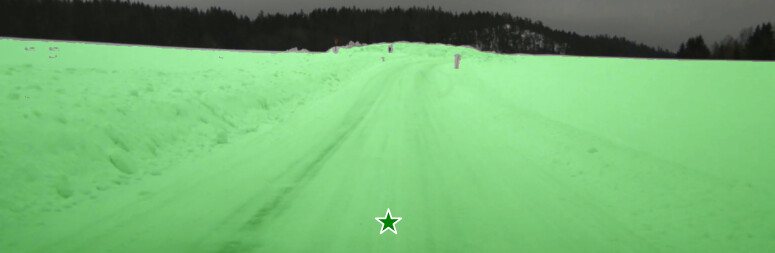} &
\includegraphics[width = 1.4in]{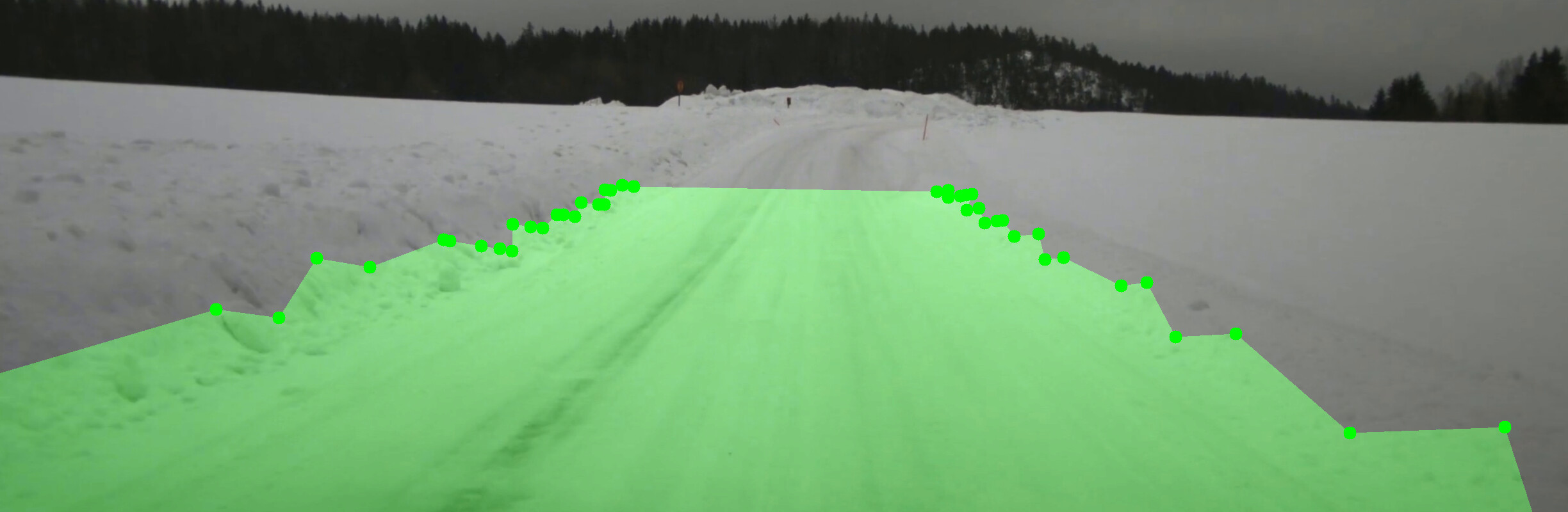} &
\includegraphics[width = 1.4in]{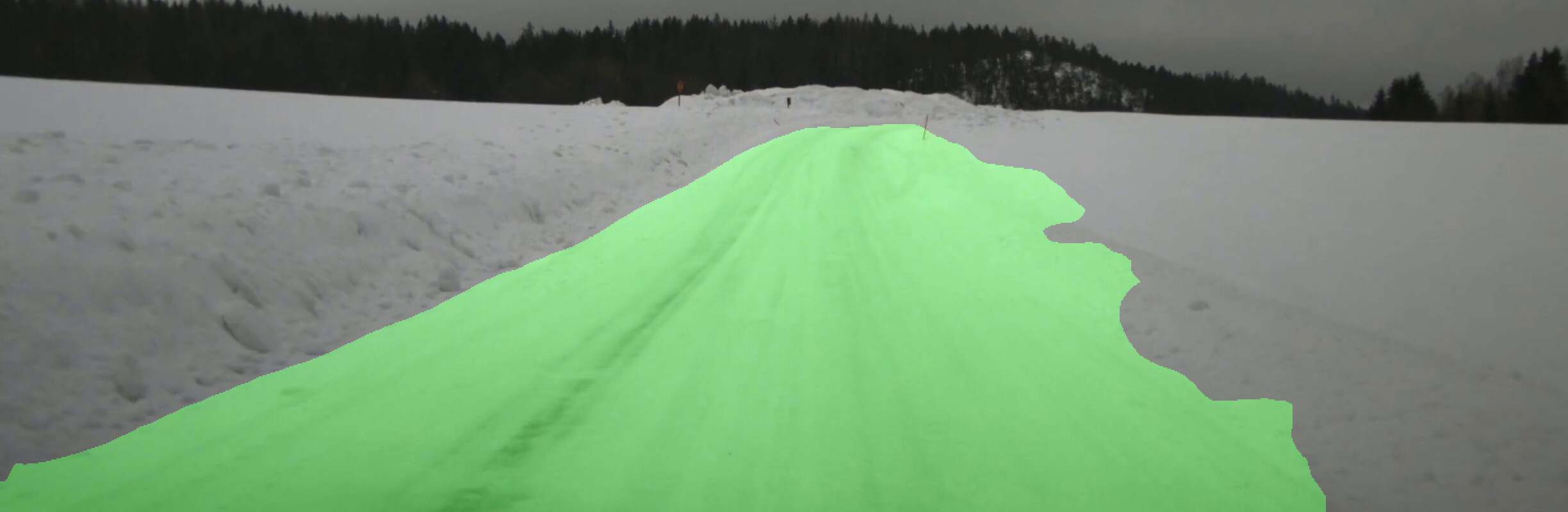} &
\includegraphics[width = 1.4in]{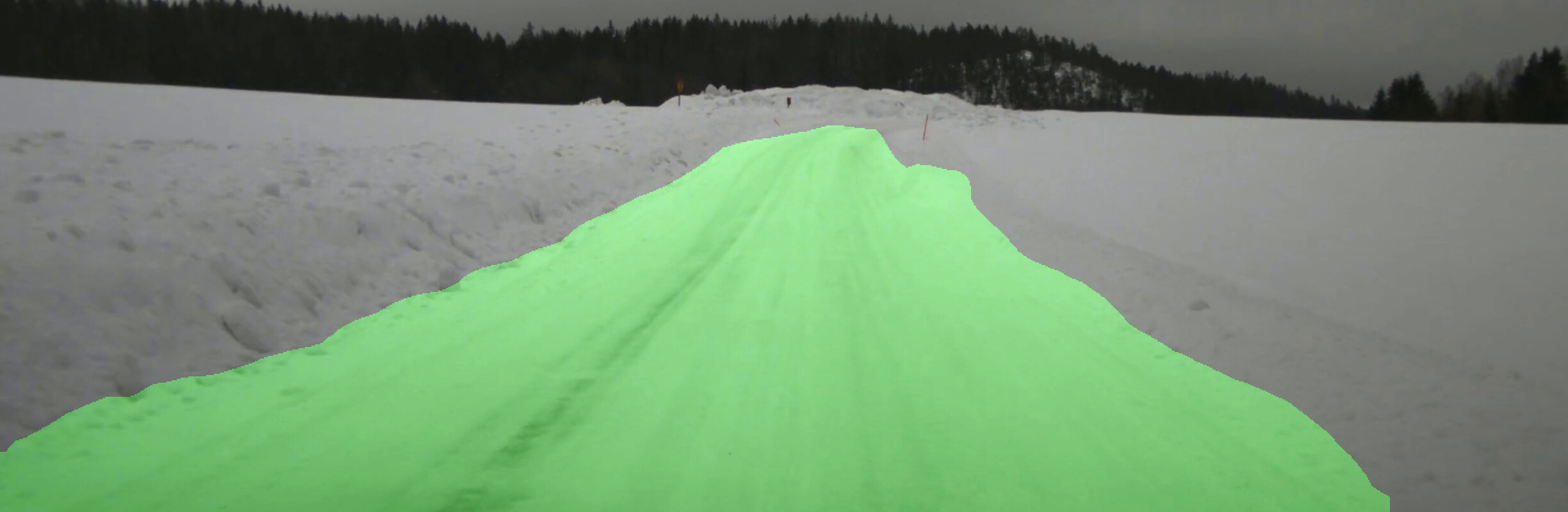} \\

\includegraphics[width = 1.4in]{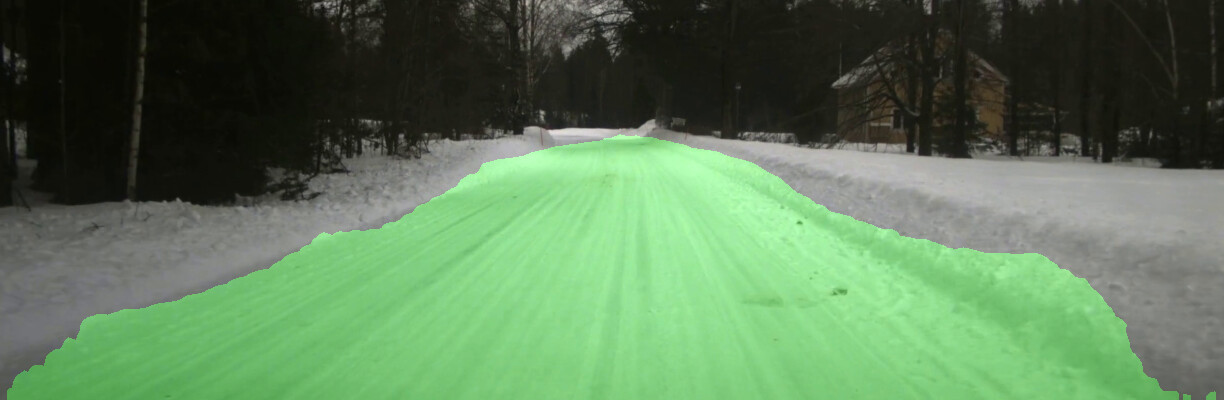} &
\includegraphics[width = 1.4in]{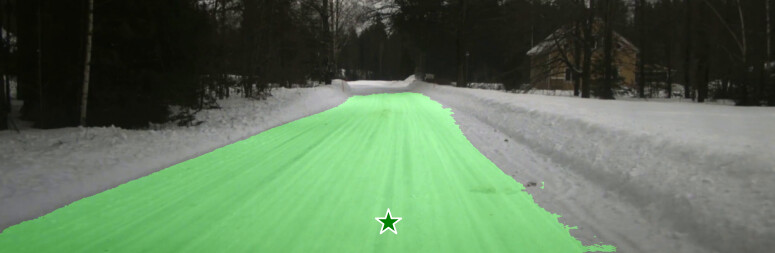} &
\includegraphics[width = 1.4in]{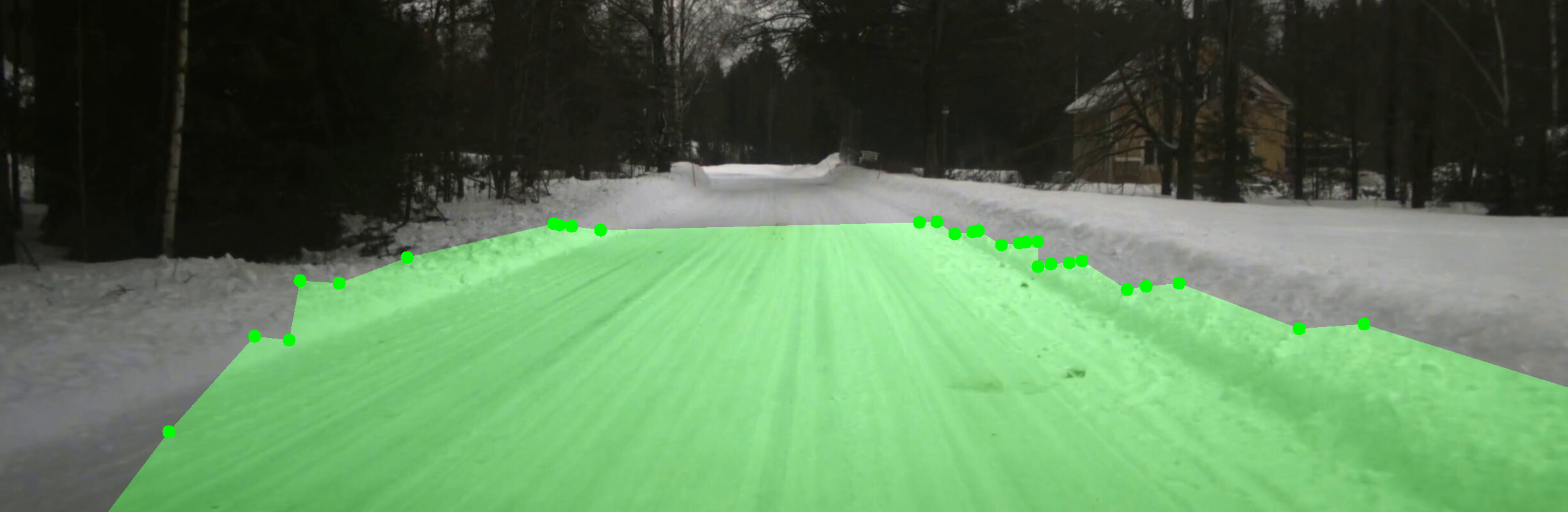} &
\includegraphics[width = 1.4in]{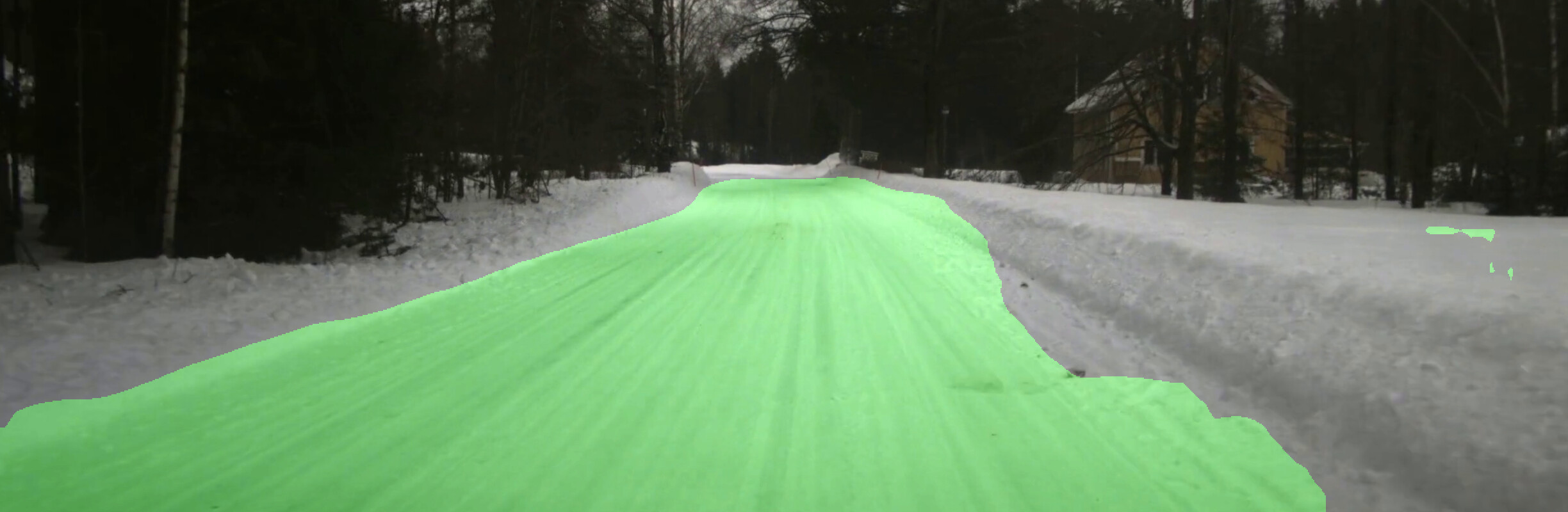} &
\includegraphics[width = 1.4in]{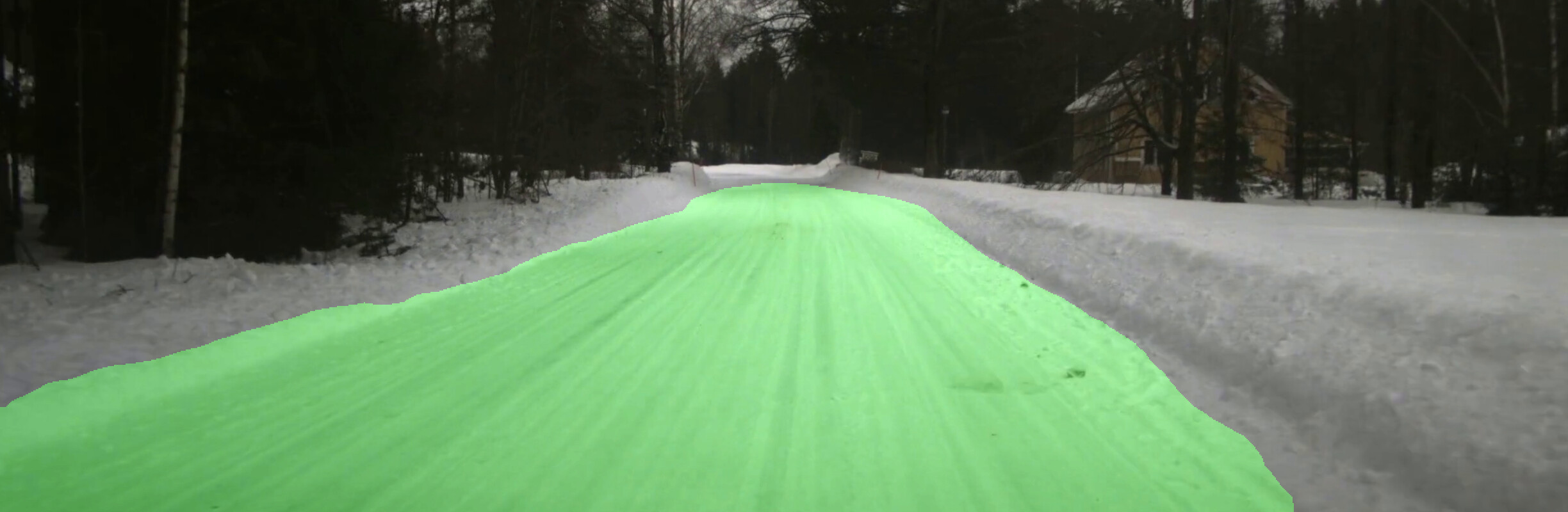} \\

\end{tabular}
\caption{Our predictions compared to recent baselines.}
\label{fig:preds}
\end{figure*}

Our autolabels were further evaluated by training a prediction model on them. 
We chose Deeplabv3 \cite{chen2017rethinking} for this purpose, referring to the Deeplabv3 trained with our autolabels as OURS.
While lidar data is used in autolabeling, our prediction model is fully image-based.
We compare our model's performance against several baselines (Table \ref{tab:preds}): a zero-shot segmentation model (SAM2 \cite{ravi2024sam}), a lidar boundary detection model \cite{wang2020speed}, a trajectory-based traversability model (Seo et. al.\cite{seo2023learning}), and a supervised model.
We also present examples of predictions for each method (Fig. \ref{fig:preds}).

In the tests, our model achieves the highest IoU and F1-score. We also have the highest precision and second-highest recall. 
Seo struggles with distinguishing road from background in visually similar areas. 
SAM2 underestimates the road segment in suburban scenes but overestimates roadside areas as drivable in the countryside scene. 
Lidar boundary detection works when boundaries are clear but fails in unstructured environments. 
The supervised model has a performance close to ours but struggles more with indistinct countryside road boundaries. 

To illustrate the advantage of separating the autolabeling process from the prediction model, we also report runtime. 
While our autolabeling process takes a few seconds per frame, a fast prediction model can be trained with this data. 
Our Deeplabv3 model runs inference in under 3 ms, allowing real-time deployment. 

\begin{table*}[]
    \centering
    \caption{Ablation study of the proposed autolabeling components for our dataset. 
    Intersection over Union (IoU), Precision (PRE), Recall (REC), and F1-score (F1) are reported. 
    Highest value in \textbf{bold}.}
    \begin{tabular}{|l | c | c | c|}
        \hline
        & All & Countryside scene & Suburb scene \\
        Configuration & IoU \ PRE \ REC \ F1 & IoU \ PRE \ REC \ F1 & IoU \ PRE \ REC \ F1 \\
        \hline
        Camera-based (C) & 83.5 \ 92.9 \ 89.2 \ 91.0 & 86.5 \ 94.6 \ 91.0 \ 92.8 & 81.0 \ 91.5 \ 87.7 \ 89.5  \\
        Height-based (H) & 80.2 \ 81.8 \ 97.7 \ 89.0 & 74.4 \ 75.1 \ 98.8 \ 85.3 & 86.1 \ 88.7 \ 96.7 \ 92.5 \\
        Gradient-based no thresholding & 34.0 \ \textbf{97.6} \ 34.2 \ 50.7 & 39.0 \ \textbf{96.5} \ 39.6 \ 56.1 & 29.6 \ \textbf{98.9} \ 29.7 \ 45.7 \\
        Gradient-based (G) & 80.5 \ 93.5 \ 85.3 \ 89.2 & 82.8 \ 91.8 \ 89.4 \ 90.6 & 78.5 \ 95.2 \ 81.8 \ 88.0 \\
        H + G & 87.0 \ 92.1 \ 94.0 \ 93.0 & 86.4 \ 88.5 \ 97.3 \ 92.7 & 87.5 \ 95.5 \ 91.3 \ 93.4 \\
        C + H & 83.3 \ 84.8 \ \textbf{97.8} \ 90.9 & 78.5 \ 79.1 \ \textbf{99.0} \ 88.0 & 88.0 \ 90.6 \ \textbf{96.8} \ 93.6  \\
        C + G & 84.3 \ 95.1 \ 88.2 \ 91.5 & 86.6 \ 93.9 \ 91.7 \ 92.8 & 82.4 \ 96.2 \ 85.1 \ 90.3 \\
        C + H + G & 89.8 \ 94.5 \ 94.8 \ 94.6 & 90.3 \ 92.5 \ 97.5 \ 94.9 & 89.4 \ 96.3 \ 92.5 \ 94.4 \\
        \textit{C + H + G + CRF} & \textbf{90.2} \ 94.6 \ 95.0 \ \textbf{94.8} & \textbf{90.5} \ 92.6 \ 97.5 \ \textbf{95.0} & \textbf{89.9} \ 96.5 \ 92.9 \ \textbf{94.7} \\
        \hline
    \end{tabular}
    \label{tab:ablations}
\end{table*}

\textbf{}

\begin{table*}[]
    \centering
    \caption{Prediction performance. 
    Intersection over Union (IoU), Precision (PRE), Recall (REC), and F1-score (F1) are reported. 
    Highest value in \textbf{bold}. 
    Tested on Nvidia RTX 3070.}
    \begin{tabular}{|l | c | c | c | c|}
        \hline
        & All & Countryside scene & Suburb scene &  \\
        Method & IoU \ PRE \ REC \ F1 & IoU \ PRE \ REC \ F1 & IoU \ PRE \ REC \ F1 & Runtime (ms) \\
        \hline
        Seo et. al. & 75.2 \ 81.3 \ 90.9 \ 85.8 & 72.3 \ 74.1 \ 96.8 \ 83.9 & 78.1 \ 89.7 \ 85.8 \ 87.7 & 4.9  \\
        SAM 2 & 67.3 \ 79.1 \ 82.0 \ 80.5 & 65.0 \ 67.1 \ 95.4 \ 78.8 & 70.3 \ \textbf{99.6} \ 70.5 \ 82.6 & 120.6 \\
        Lidar Boundary Detection & 70.2 \ 74.5 \ 92.4 \ 82.5 & 68.8 \ 72.2 \ 93.6 \ 81.5 & 71.4 \ 76.6 \ 91.4 \ 83.3 & 3.8 \\
        Supervised & 89.7 \ 93.5 \ 95.7 \ 94.6 & 88.8 \ 91.8 \ 96.5 \ 94.1 & 90.5 \ 95.0 \ \textbf{95.0} \ 95.0 & 2.7 \\
        \textit{OURS} & \textbf{90.9} \ \textbf{94.5} \ \textbf{95.9} \ \textbf{95.2} & \textbf{91.1} \ \textbf{92.8} \ \textbf{97.9} \ \textbf{95.3} & \textbf{90.7} \ 96.1 \ 94.2 \ \textbf{95.1} & 2.7 \\
        \hline
    \end{tabular}
    \label{tab:preds}
\end{table*}

\section{Discussion}

\begin{figure}
    \centering
    \includegraphics[width=\linewidth]{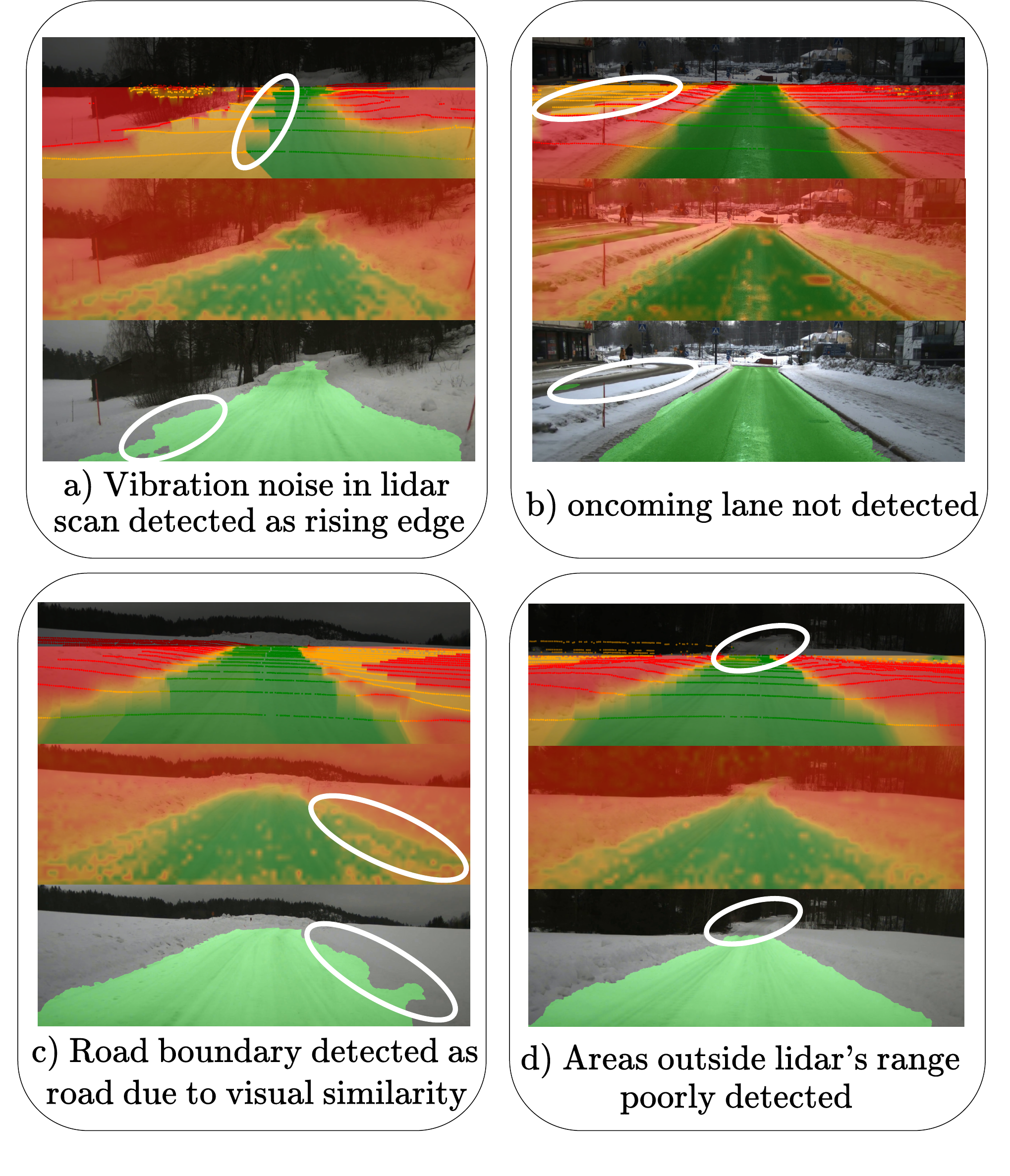}
    \caption{Failure cases of the proposed autolabeling method.}
    \label{fig:failure_cases}
\end{figure}

This paper introduced a novel trajectory-based lidar-camera fusion method for automatic road labeling, outperforming existing methods on our winter driving dataset. 
The fusion method provides increased reliability by including two independent autolabeling methods. 
Most common failure cases are presented in Fig \ref{fig:failure_cases}.   
In case a) vibration causes noise to the lidar scan which is detected as a high gradient by the lidar-autolabeling. Good camera-based detection decreases the error in the final label. 
In case b) camera-autolabeling detects road boundary as road due to visual similarity. Good lidar-based detection decreases the error in the final label. 
In case c) oncoming lane is not detected by the lidar-autolabeling due to the upward gradients caused by the snowbank. The gradient-based autolabeling can't handle two separate road segments. 
In case d) road outside lidar's range is poorly detected. We rely on a single scan and image limiting long-distance labeling capabilities. In future work consecutive detections could be accumulated for an extended range. 

Currently, our lidar-based autolabeling method assumes that the road elevation is lower than its surroundings and only considers upward height changes for detecting roadside areas. 
In structured environments and winter driving this assumption usually holds, but for off-road adaption also downward elevation changes should be considered. We also assume that the lidar scans aren't badly degraded. Lidar scan quality can be compromised for example by water layer on top of asphalt. Such scenarios were not included in the study. 

\section{Acknowledgements}
We express our gratitude to the Aalto Doctoral School Program and Henry Ford Foundation Finland for providing funding for this research. We acknowledge the computational resources provided by the Aalto Science-IT project.

\bibliographystyle{ieeetr}

\bibliography{ref}

\begin{thebibliography}{10}

\bibitem{geiger2013vision}
A.~Geiger, P.~Lenz, C.~Stiller, and R.~Urtasun, ``Vision meets robotics: The kitti dataset,'' {\em The International Journal of Robotics Research}, vol.~32, no.~11, pp.~1231--1237, 2013.

\bibitem{caesar2020nuscenes}
H.~Caesar, V.~Bankiti, A.~H. Lang, S.~Vora, V.~E. Liong, Q.~Xu, A.~Krishnan, Y.~Pan, G.~Baldan, and O.~Beijbom, ``nuscenes: A multimodal dataset for autonomous driving,'' in {\em Proceedings of the IEEE/CVF conference on computer vision and pattern recognition}, pp.~11621--11631, 2020.

\bibitem{yu2020bdd100k}
F.~Yu, H.~Chen, X.~Wang, W.~Xian, Y.~Chen, F.~Liu, V.~Madhavan, and T.~Darrell, ``Bdd100k: A diverse driving dataset for heterogeneous multitask learning,'' in {\em Proceedings of the IEEE/CVF conference on computer vision and pattern recognition}, pp.~2636--2645, 2020.

\bibitem{cordts2016cityscapes}
M.~Cordts, M.~Omran, S.~Ramos, T.~Rehfeld, M.~Enzweiler, R.~Benenson, U.~Franke, S.~Roth, and B.~Schiele, ``The cityscapes dataset for semantic urban scene understanding,'' in {\em Proceedings of the IEEE conference on computer vision and pattern recognition}, pp.~3213--3223, 2016.

\bibitem{shaik2024idd}
F.~A. Shaik, A.~Reddy, N.~R. Billa, K.~Chaudhary, S.~Manchanda, and G.~Varma, ``Idd-aw: A benchmark for safe and robust segmentation of drive scenes in unstructured traffic and adverse weather,'' in {\em Proceedings of the IEEE/CVF Winter Conference on Applications of Computer Vision}, pp.~4614--4623, 2024.

\bibitem{sakaridis2021acdc}
C.~Sakaridis, D.~Dai, and L.~Van~Gool, ``Acdc: The adverse conditions dataset with correspondences for semantic driving scene understanding,'' in {\em Proceedings of the IEEE/CVF International Conference on Computer Vision}, pp.~10765--10775, 2021.

\bibitem{schmid2022self}
R.~Schmid, D.~Atha, F.~Sch{\"o}ller, S.~Dey, S.~Fakoorian, K.~Otsu, B.~Ridge, M.~Bjelonic, L.~Wellhausen, M.~Hutter, {\em et~al.}, ``Self-supervised traversability prediction by learning to reconstruct safe terrain,'' in {\em 2022 IEEE/RSJ International Conference on Intelligent Robots and Systems (IROS)}, pp.~12419--12425, IEEE, 2022.

\bibitem{seo2023learning}
J.~Seo, S.~Sim, and I.~Shim, ``Learning off-road terrain traversability with self-supervisions only,'' {\em IEEE Robotics and Automation Letters}, vol.~8, no.~8, pp.~4617--4624, 2023.

\bibitem{jung2024v}
S.~Jung, J.~Lee, X.~Meng, B.~Boots, and A.~Lambert, ``V-strong: Visual self-supervised traversability learning for off-road navigation,'' in {\em 2024 IEEE International Conference on Robotics and Automation (ICRA)}, pp.~1766--1773, IEEE, 2024.

\bibitem{seo2023scate}
J.~Seo, T.~Kim, K.~Kwak, J.~Min, and I.~Shim, ``Scate: A scalable framework for self-supervised traversability estimation in unstructured environments,'' {\em IEEE Robotics and Automation Letters}, vol.~8, no.~2, pp.~888--895, 2023.

\bibitem{alamikkotervo2024tadap}
E.~Alamikkotervo, R.~Ojala, A.~Sepp{\"a}nen, and K.~Tammi, ``Tadap: Trajectory-aided drivable area auto-labeling with pretrained self-supervised features in winter driving conditions,'' {\em IEEE Transactions on Intelligent Vehicles}, 2024.

\bibitem{liu2018co}
Z.~Liu, S.~Yu, and N.~Zheng, ``A co-point mapping-based approach to drivable area detection for self-driving cars,'' {\em Engineering}, vol.~4, no.~4, pp.~479--490, 2018.

\bibitem{sock2016probabilistic}
J.~Sock, J.~Kim, J.~Min, and K.~Kwak, ``Probabilistic traversability map generation using 3d-lidar and camera,'' in {\em 2016 IEEE international conference on robotics and automation (ICRA)}, pp.~5631--5637, IEEE, 2016.

\bibitem{frey2024roadrunner}
J.~Frey, S.~Khattak, M.~Patel, D.~Atha, J.~Nubert, C.~Padgett, M.~Hutter, and P.~Spieler, ``Roadrunner--learning traversability estimation for autonomous off-road driving,'' {\em arXiv preprint arXiv:2402.19341}, 2024.

\bibitem{chen2023learning}
E.~Chen, C.~Ho, M.~Maulimov, C.~Wang, and S.~Scherer, ``Learning-on-the-drive: Self-supervised adaptation of visual offroad traversability models,'' {\em arXiv preprint arXiv:2306.15226}, 2023.

\bibitem{wang2019self}
H.~Wang, Y.~Sun, and M.~Liu, ``Self-supervised drivable area and road anomaly segmentation using rgb-d data for robotic wheelchairs,'' {\em IEEE Robotics and Automation Letters}, vol.~4, no.~4, pp.~4386--4393, 2019.

\bibitem{mayr2018self}
J.~Mayr, C.~Unger, and F.~Tombari, ``Self-supervised learning of the drivable area for autonomous vehicles,'' in {\em 2018 IEEE/RSJ International Conference on Intelligent Robots and Systems (IROS)}, pp.~362--369, IEEE, 2018.

\bibitem{ma2023self}
F.~Ma, Y.~Liu, S.~Wang, J.~Wu, W.~Qi, and M.~Liu, ``Self-supervised drivable area segmentation using lidar's depth information for autonomous driving,'' in {\em 2023 IEEE/RSJ International Conference on Intelligent Robots and Systems (IROS)}, pp.~41--48, IEEE, 2023.

\bibitem{rawashdeh2023camera}
N.~A. Rawashdeh, J.~P. Bos, and N.~J. Abu-Alrub, ``Camera--lidar sensor fusion for drivable area detection in winter weather using convolutional neural networks,'' {\em Optical Engineering}, vol.~62, no.~3, pp.~031202--031202, 2023.

\bibitem{caron2021emerging}
M.~Caron, H.~Touvron, I.~Misra, H.~J{\'e}gou, J.~Mairal, P.~Bojanowski, and A.~Joulin, ``Emerging properties in self-supervised vision transformers,'' in {\em Proceedings of the IEEE/CVF international conference on computer vision}, pp.~9650--9660, 2021.

\bibitem{oquab2023dinov2}
M.~Oquab, T.~Darcet, T.~Moutakanni, H.~Vo, M.~Szafraniec, V.~Khalidov, P.~Fernandez, D.~Haziza, F.~Massa, A.~El-Nouby, {\em et~al.}, ``Dinov2: Learning robust visual features without supervision,'' {\em arXiv preprint arXiv:2304.07193}, 2023.

\bibitem{kirillov2023segment}
A.~Kirillov, E.~Mintun, N.~Ravi, H.~Mao, C.~Rolland, L.~Gustafson, T.~Xiao, S.~Whitehead, A.~C. Berg, W.-Y. Lo, {\em et~al.}, ``Segment anything,'' in {\em Proceedings of the IEEE/CVF International Conference on Computer Vision}, pp.~4015--4026, 2023.

\bibitem{ravi2024sam}
N.~Ravi, V.~Gabeur, Y.-T. Hu, R.~Hu, C.~Ryali, T.~Ma, H.~Khedr, R.~R{\"a}dle, C.~Rolland, L.~Gustafson, {\em et~al.}, ``Sam 2: Segment anything in images and videos,'' {\em arXiv preprint arXiv:2408.00714}, 2024.

\bibitem{sun20193d}
P.~Sun, X.~Zhao, Z.~Xu, R.~Wang, and H.~Min, ``A 3d lidar data-based dedicated road boundary detection algorithm for autonomous vehicles,'' {\em IEEE Access}, vol.~7, pp.~29623--29638, 2019.

\bibitem{wang2020speed}
G.~Wang, J.~Wu, R.~He, and B.~Tian, ``Speed and accuracy tradeoff for lidar data based road boundary detection,'' {\em IEEE/CAA Journal of Automatica Sinica}, vol.~8, no.~6, pp.~1210--1220, 2020.

\bibitem{zhang2018road}
Y.~Zhang, J.~Wang, X.~Wang, and J.~M. Dolan, ``Road-segmentation-based curb detection method for self-driving via a 3d-lidar sensor,'' {\em IEEE transactions on intelligent transportation systems}, vol.~19, no.~12, pp.~3981--3991, 2018.

\bibitem{krahenbuhl2011efficient}
P.~Kr{\"a}henb{\"u}hl and V.~Koltun, ``Efficient inference in fully connected crfs with gaussian edge potentials,'' {\em Advances in neural information processing systems}, vol.~24, 2011.

\bibitem{chen2017rethinking}
L.-C. Chen, ``Rethinking atrous convolution for semantic image segmentation,'' {\em arXiv preprint arXiv:1706.05587}, 2017.

\end{thebibliography}


\end{document}